\PassOptionsToPackage{dvipsnames}{xcolor}
\documentclass[11pt,a4paper]{academicreport}

\usepackage{longtable}
\usepackage{tabularray}
\usepackage{wrapfig}
\usepackage{placeins}
\usepackage{makecell}
\usepackage{tablefootnote}
\usepackage{threeparttable}
\usepackage{arydshln}
\usepackage{nicefrac}
\usepackage{bm}
\usepackage{inconsolata}
\usepackage{fontawesome5}

\colorlet{goldaccent}{accent}
\colorlet{golddark}{accent}
\colorlet{goldfg}{black}
\colorlet{goldbg}{accentlight}

\newcolumntype{C}[1]{>{\centering\arraybackslash}m{#1}}

\theoremstyle{definition}

\newcommand{\paratitle}[1]{\paragraph{#1}}

\newcommand{\ourmethod}{iCoder\xspace}

\newcommand{\tableicon}[1]{
  \makebox[1.15em][c]{\faIcon{#1}}\hspace{0.25em}
}
\newcommand{\skillrow}[2]{
  \multicolumn{3}{@{}p{0.92\linewidth}@{}}{
    \begingroup
    \setlength{\fboxsep}{2.5pt}
    \colorbox{accent!4}{
      \parbox{\dimexpr\linewidth-2\fboxsep\relax}{
        {\normalfont\scriptsize\bfseries\color{accent}#1}\\[0.5pt]
        {\ttfamily\scriptsize #2}
      }
    }
    \endgroup
  } \\
}

\providecommand{\bibfont}{\small}

\usepackage{fancyhdr}
\title{
  \fontsize{15pt}{18pt}\selectfont
  \includegraphics[width=0.36\textwidth]{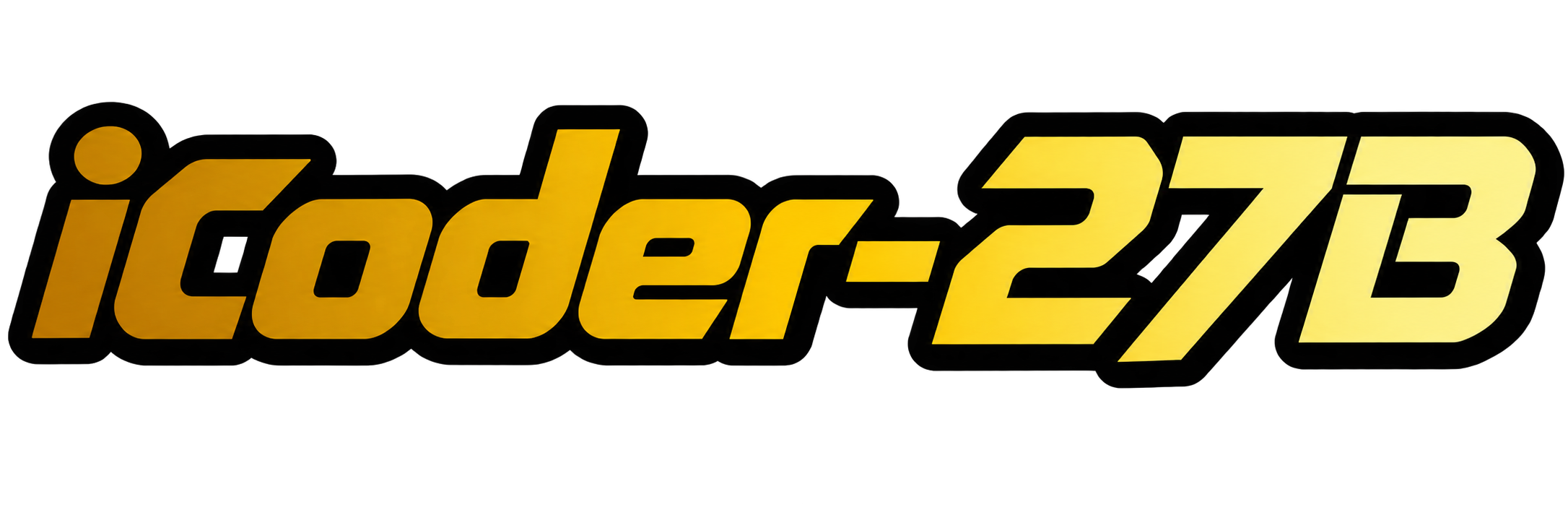}\\[-0.95em]
  \textbf{Recursive AI-Led Development of Frontier
  \textcolor{accent}{Industrial Coding Model}}
}
\author{
  \small
  \textbf{Core Contributors:}
  Cheng Yang\textsuperscript{\ding{68}},
  Jiayang Lyu\textsuperscript{\ding{68}},
  Shangyuan Liu\textsuperscript{\ding{68}};\quad
  \textbf{Contributors:}
  Jiong Lin\textsuperscript{\ding{68}},
  Xinlei Yu\textsuperscript{\ding{75}}\\[0.3em]
  \textbf{Project Lead:}
  Guibin Zhang\textsuperscript{\ding{75}};\quad
  \textbf{Senior Advisors:}
  Junchi Yan\textsuperscript{\ding{68}},
  Shuicheng Yan\textsuperscript{\ding{75}}, Weinan E\textsuperscript{\ding{68}},\\[0.3em]
  \textbf{Corresponding Authors:}
  Linfeng Zhang\textsuperscript{\ding{68}},
  Linfeng Zhang\textsuperscript{\ding{168}\,\ding{161}},
  Qibing Ren\textsuperscript{\ding{68}}\\[0.4em]
  \textsuperscript{\ding{68}}School of Artificial Intelligence, Shanghai Jiao Tong University\\
  \textsuperscript{\ding{168}}DP Technology \quad
  \textsuperscript{\ding{75}}National University of Singapore \quad
  \textsuperscript{\ding{161}}Endless Frontier
}

\date{}

\fancypagestyle{firstpage}{
  \fancyhf{}
  \fancyhead[L]{
    \includegraphics[height=26pt]{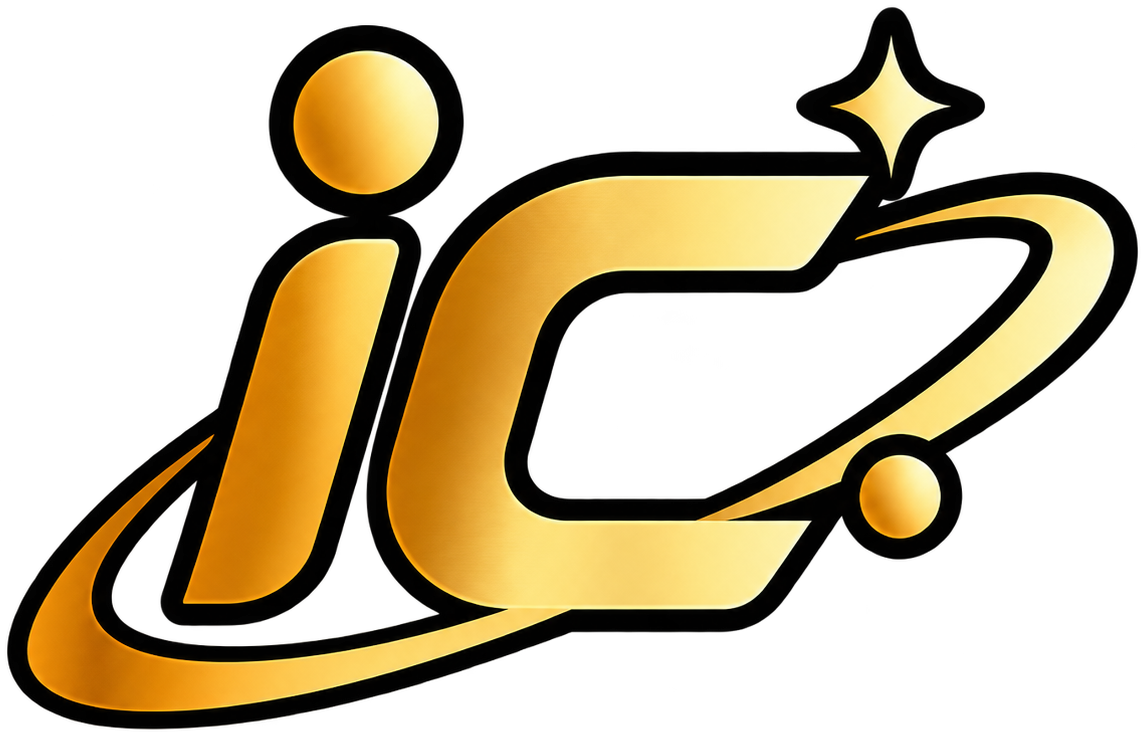}
  }
  \fancyhead[R]{
    \includegraphics[height=26pt]{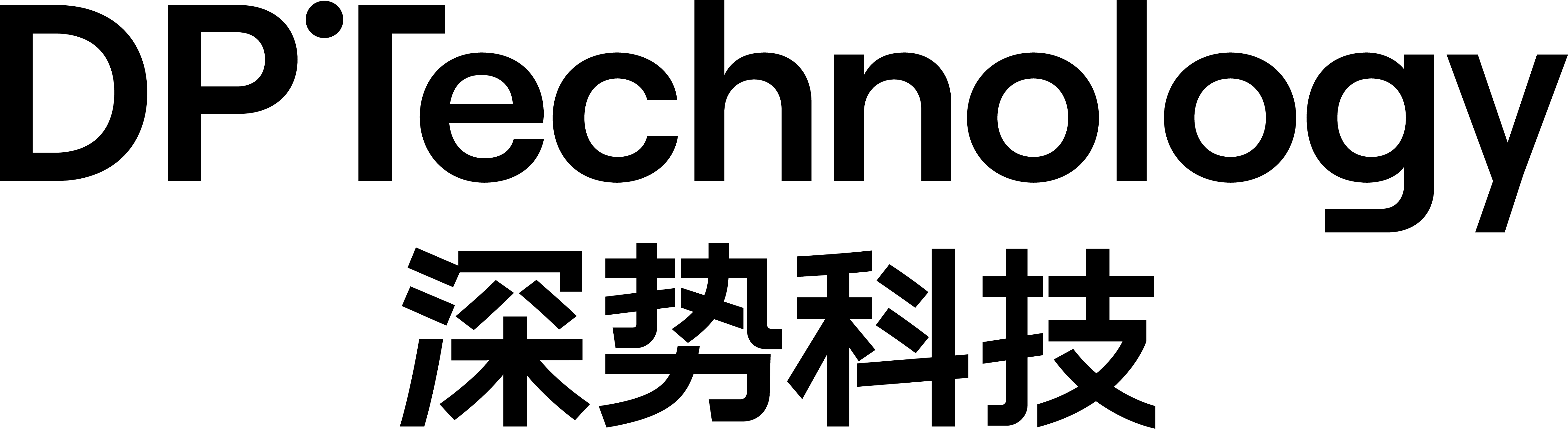}\hspace{0.9em}
    \includegraphics[height=26pt]{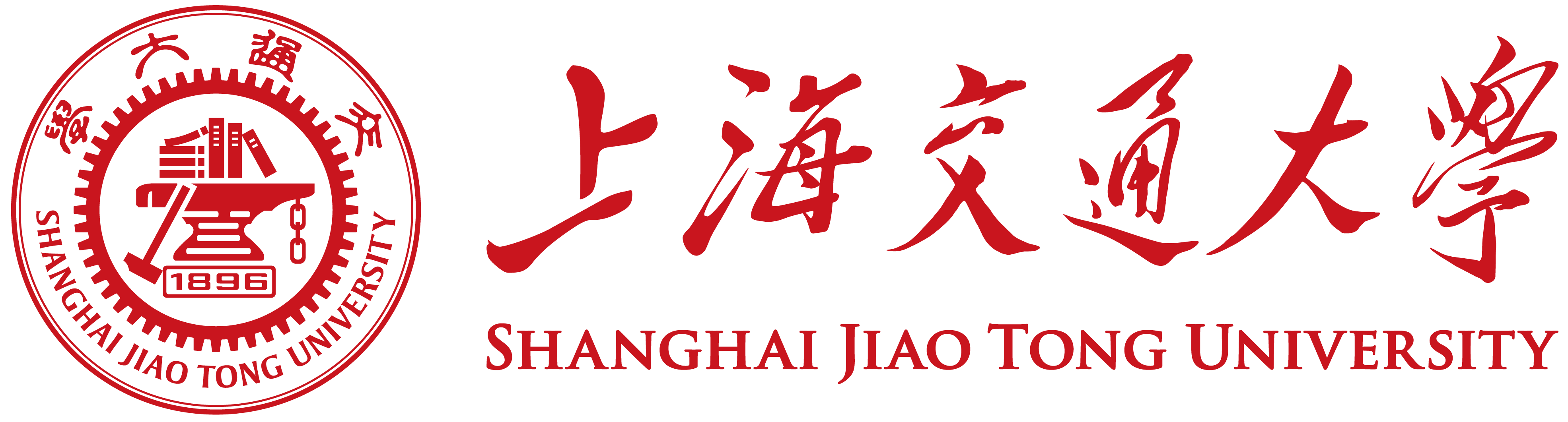}\hspace{0.9em}
    \includegraphics[height=26pt]{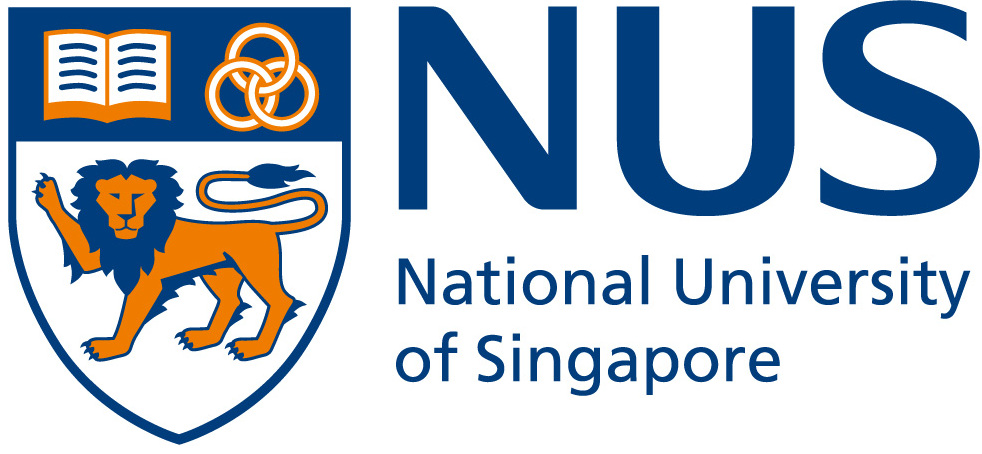}
  }
  \fancyfoot[C]{\thepage}
  
}

\begin{document}

\maketitle
\thispagestyle{firstpage}
\vspace{-0.6em}
\begin{abstract}
Recursive AI, the prospect of AI taking an increasingly complete role in
building and improving AI, is a crown jewel of AI for AI. Although recursive
self-development has become practical for small models, bounded tasks, and fixed
time budgets, a more consequential realization of this ambition, i.e., developing a release-ready, frontier-competitive model, remains far
more challenging.
In this work,
we ask how little human involvement is sufficient for an agent to develop a
frontier model. We concentrate human input into a high-density, low-frequency
interface: experts encode objectives, stage scaffolds, permission boundaries,
and operating procedures as reusable research skills, while the agent
instantiates these priors, selects experiments, diagnoses outcomes, and revises
the training strategy. In the challenging domain of industrial coding, the
agent evolves data and coordinates SFT, on-policy self-distillation, and
reinforcement learning with verifiable rewards, ultimately producing
\ourmethod, a 27B model for RTL design and GPU kernel optimization. Across
seven benchmarks, \ourmethod leads RTLLM, outperforming GPT-5.5
and Claude-Opus-4.8; ranks second on CVDP and KernelBench L2,
exceeding GPT-5.5 by 16 points; and ties Claude-Opus-4.8 for the
best TritonBench result. Exploratory case studies further show \ourmethod's competitive
iterative RTL and GPU-kernel optimization with substantially fewer tokens. These results chart an engineering path toward recursive
self-improvement, in which humans distill the principles of model building,
agents operationalize them through evidence-driven experimentation, and each
generation of AI becomes a more capable architect of the next.
\par\medskip
{\centering\small
\renewcommand{\arraystretch}{1.15}
\begin{tabular}{@{}c@{\hspace{0.6em}}l@{\hspace{0.6em}}l@{}}

\faGithub & \textbf{\textcolor{accent}{GitHub:}} &
  \href{https://github.com/bingreeky/iCoder}{\nolinkurl{https://github.com/bingreeky/iCoder}} \\
\raisebox{-0.15\height}{\includegraphics[height=1.05em]{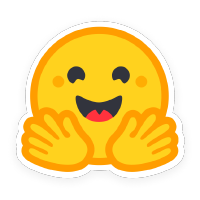}} &
  \textbf{\textcolor{accent}{Hugging Face:}} &
  \href{https://huggingface.co/i-Coder}{\nolinkurl{https://huggingface.co/i-Coder}}
\end{tabular}
\par}
\end{abstract}

\begin{figure}[!hb]
  \centering

  \includegraphics[width=\linewidth]{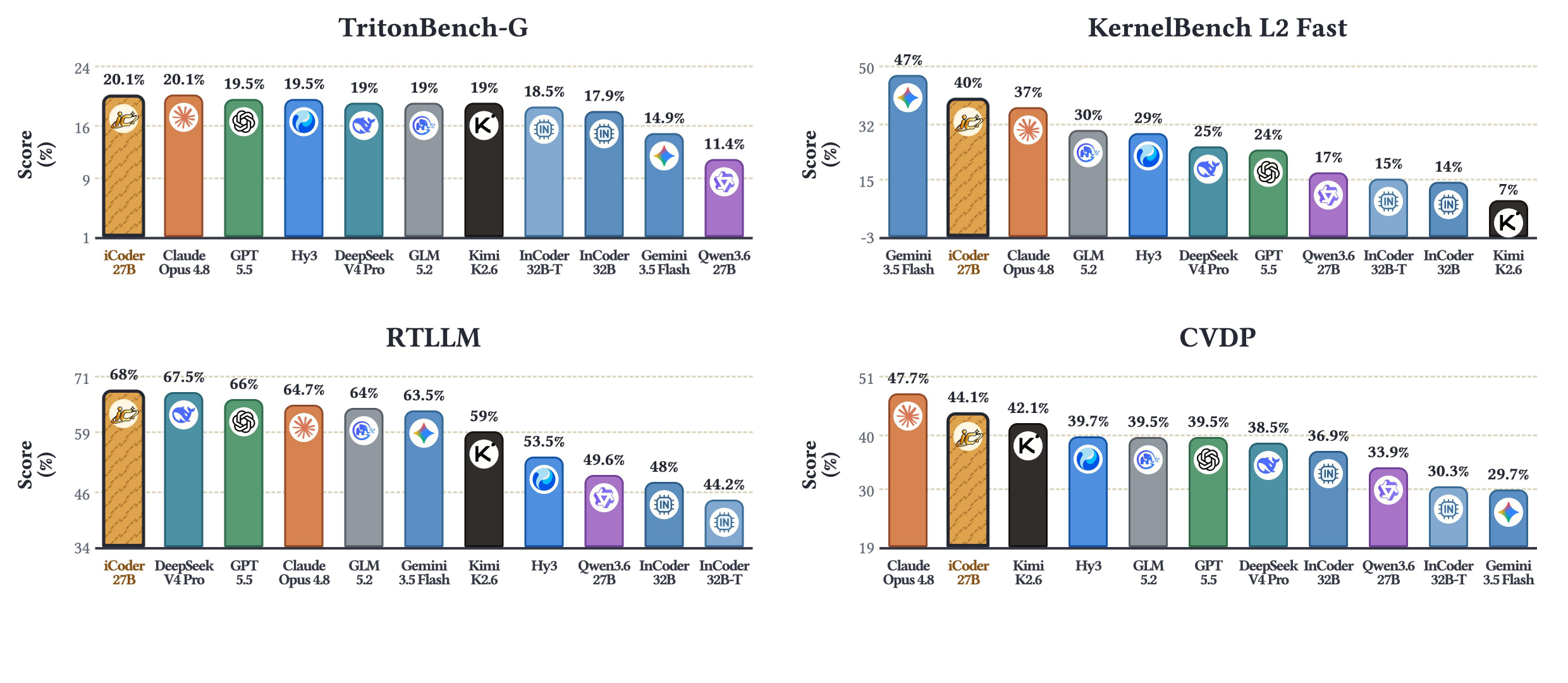}
  \vspace{-3em}
  \caption{\textbf{Industrial coding performance across kernel and RTL
  benchmarks.} We compare \ourmethod{} against leading open and proprietary
  models on TritonBench-G, KernelBench L2 Fast, RTLLM, and CVDP. Bars are ordered
  independently within each benchmark, and higher scores are better. Despite
  its compact 27B scale, \ourmethod{} ranks second on KernelBench L2 Fast and
  CVDP, leads on RTLLM, and ties the best result on TritonBench-G.}
  \label{fig:industrial-benchmark}
  \vspace{-0.85em}
\end{figure}

\section{Introduction}
\label{sec:introduction}

AI agents are moving from assisting isolated research tasks to executing
increasingly complete scientific workflows \citep{yamada2026endtoend,zhao2026scienceflow}.
Recent systems can carry an AI project from hypothesis formation through
experimentation and manuscript production \citep{yamada2026endtoend}, search
executable solution spaces under quantitative feedback \citep{lin2026phynex},
and sustain long-horizon investigation through persistent state and adaptive
compute allocation \citep{zhao2026scienceflow}. These advances have renewed
interest in recursive self-improvement (RSI), yet the strongest evidence still
comes from bounded loops grounded by externally supplied evaluators
\citep{chen2026rsi,ding2026verificationgap}. The following question remains challenging:

\begin{insightbox}{Research Question}
\emph{Can AI coordinate the development of a genuinely competitive
frontier model?}
\end{insightbox}

\begin{figure}[!t]
  \centering
  \includegraphics[width=\linewidth]{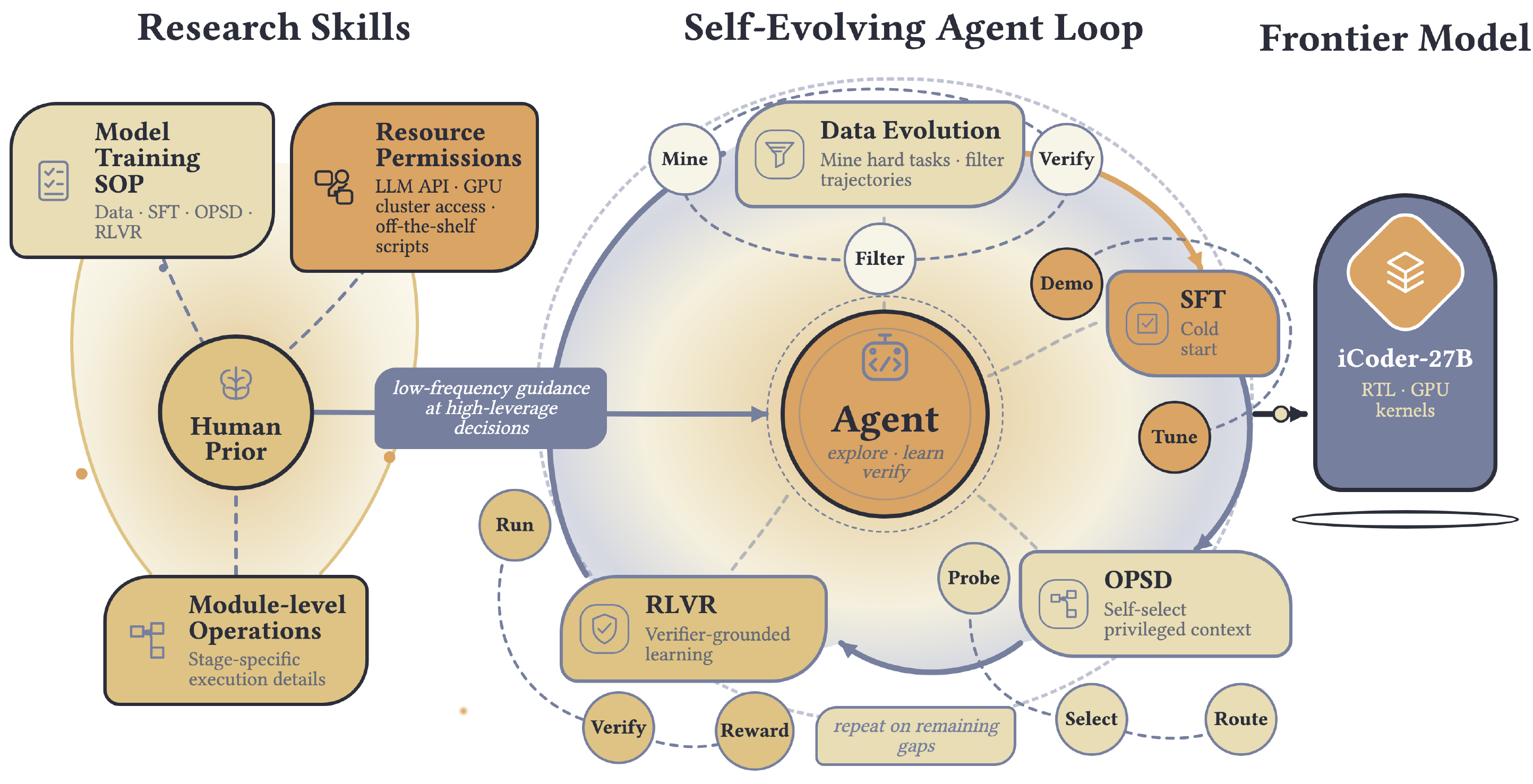}
  \caption{\textbf{Overview of our human-guided self-evolving training
  framework.} Research skills are distilled into model-training SOPs, resource
  permissions, and module-level operations, providing low-frequency human
  guidance at high-leverage decisions. The agent then autonomously iterates
  through data evolution, SFT, OPSD, and RLVR to close remaining capability
  gaps and produce the frontier model, iCoder-27B.}
  \label{fig:intro-pipeline}
\end{figure}

Frontier-model development sets a qualitatively higher bar than improving one
program, harness, or benchmark score. Recent RSI systems iterate within sealed
sandboxes whose data, evaluator, and action space are fixed
\citep{guo2026aqua}, or refine prompt-level harnesses over synthetic research
tasks \citep{lee2026rhi}. Even a long-running study spanning roughly one hundred
experiments remains confined to a single architecture problem and relies on
human-declared phase transitions \citep{safdar2026longhorizon}. Developing a
release-ready foundation model, by contrast, requires a tightly coupled stack:
choices made while constructing data reshape distributed training, post-training,
evaluation, and ultimately the next round of decisions
\citep{qwen2026qwen36,deepseek2026v4,li2026trex}. The experiments are expensive,
failures propagate across stages, and their causes are rarely localized. The
central question is thus not simply whether an agent can optimize, but what
expert knowledge it must receive, at what granularity, and how often humans must
re-enter the loop.

We do not claim to leap directly to fully closed RSI. Instead, we study a
practical intermediate target: how little human involvement is sufficient for
an agent to lead the development of a release-ready, frontier-competitive model?
Our answer is \emph{high-density prior, low-frequency intervention}. We encode
expert model-development experience once as human-authored \emph{Research
Skills}: concise operational procedures, backed by reusable code, that make
established practice executable on a GPU cluster. These skills turn experience
in data preparation and distributed SFT, OPSD, and RLVR into usable starting
conditions without prescribing the experiments that follow. From that point
onward, the agent chooses how to instantiate the skills, allocates resources,
runs and diagnoses experiments, and decides what to try next. The resulting
model is therefore jointly contributed by humans and the agent, but with a
deliberate boundary: humans supply the initial executable expertise, whereas
concrete model-development decisions and execution belong to the agent. We
study this division of labor in industrial coding, spanning RTL design and GPU
kernel optimization, where success demands both exact functional reasoning and
performance-aware systems optimization under executable verification.

With the research skills in place, we handed the development loop to Codex
GPT-5.6-Sol operating at xhigh reasoning effort. Starting from the base
checkpoint and available compute, the agent \ding{182} constructed an executable task pool
and \ding{183} cold-started Qwen3.6-27B on verified capability-gap trajectories before
using OPSD as a bridge to RLVR.
During \ding{184} OPSD, evaluation changed which data entered training, controlled
ablations changed what evidence the self-teacher could access, and a collapsing
objective forced the agent to redesign the learning rule around verifier-anchored
credit. During \ding{185} RLVR, reward exploits, false verdicts, and length-dependent
degeneration successively changed reward eligibility, optimization, and
trajectory budgeting. Rather than executing a fixed human-authored recipe, the
agent arrived at the final recipe through these failures and revisions. The
process ultimately produced a 27B industrial coding model, which we call
\ourmethod-27B (\ourmethod).

As summarized in Figure~\ref{fig:industrial-benchmark}, despite its compact
size, \ourmethod is highly competitive with both much
larger open models and frontier proprietary systems. Against the same-size
Qwen3.6-27B \citep{qwen2026qwen36}, it gains 18.4 points on RTLLM
(68.0 vs.\ 49.6), 16.2 points on VerilogEval Spec-to-RTL
(86.3 vs.\ 70.1), and achieves $2.64\times$ the KernelBench L2 correctness
(74 vs.\ 28). It also surpasses the 1.6T-total/49B-active DeepSeek-V4-Pro
\citep{deepseek2026v4} on RTLLM (68.0 vs.\ 67.5) and nearly doubles its
KernelBench L1 correctness (61 vs.\ 32). More strikingly, \ourmethod outperforms GPT-5.5 \citep{openai2026gpt55}, Claude
Opus 4.8 \citep{anthropic2026opus48}, and Gemini 3.5 Flash
\citep{google2026gemini35} on both RTLLM (68.0 vs.\ 66.0/64.7/63.5) and
KernelBench L1 correctness (61 vs.\ 43/55/45). On TritonBench-G, it matches
Claude Opus 4.8 at 20.1 while exceeding GPT-5.5 and Gemini 3.5 Flash at 19.5
and 14.9, respectively.
Beyond one-shot generation, evaluation-driven case studies show that
\ourmethod comes within 1.1 and 4.5 points of HY3 and DeepSeek-V4-Pro on eight
RTL designs while using 51\% and 33\% as many reported output tokens,
respectively, and reaches the strongest archived endpoint on two of eight
selected GPU operators.

Our contributions are threefold:
\begin{itemize}[leftmargin=*]
    \item We formulate frontier-model development as a real-world test of
    agentic self-improvement and provide an explicit attribution boundary
    between initial human expertise and agent-led development.
    \item We operationalize expert knowledge as research skills and document an
    auditable experimentation trajectory in which data, objectives, reward
    semantics, and resource constraints are revised in response to evidence.
    \item We deliver \ourmethod-27B, a compact 27B model that rivals or exceeds
    substantially larger open and proprietary models across
    industrial coding benchmarks.
\end{itemize}

\section{Related Work}
\label{sec:background}

\paratitle{Recursive and self-improving AI.}
Recent self-improving AI systems are best distinguished by what is revised
across iterations: an external solution, the agent machinery surrounding a
model, or the model itself
\citep{chen2026rsi,ding2026verificationgap,yamada2026endtoend,zhao2026scienceflow,luo2026xscientist}.
At the first level, a fixed model repeatedly proposes and tests candidate
solutions under evaluator feedback; improvement accumulates in the resulting
artifact rather than in the model that produced it
\citep{novikov2025alphaevolve,assumpcao2025codeevolve,du2026mlevolve,guo2026aqua,lin2026phynex}. The recursion can
instead move into the agent harness, allowing execution traces to change how a
frozen model acts on subsequent tasks
\citep{lee2026rhi,wang2026metaskill,zhou2026hsi,alzubi2026evoskill}. This raises system-level performance but
remains bounded by the capabilities of the underlying model
\citep{lee2026rhi,wang2026metaskill,zhou2026hsi,alzubi2026evoskill}. A stronger form writes the
experience back into the model parameters by training on self-generated
trajectories and verifiable outcomes
\citep{acikgoz2025selfimproving,zhao2026opsd,hubotter2026sdpo,yang2026rlsd,rank2026posttrainbench}.
Only this last form makes the acquired capability persist independently of the
particular harness and therefore most directly approaches autonomous model
development
\citep{chen2026rsi,yamada2026endtoend,acikgoz2025selfimproving,rank2026posttrainbench}.
Evidence at this level, however, remains tightly bounded by fixed task
interfaces, short budgets, or limited problem scales
\citep{rank2026posttrainbench,du2026mlevolve,safdar2026longhorizon,guo2026aqua}:
direct agentic post-training has largely been
studied on 4B-class models, with each run limited to ten hours on a single H100
and optimized against one target benchmark \citep{rank2026posttrainbench}. Our
work investigates the same model-level objective at a different operating
scale, asking an agent to coordinate a multi-stage process whose final artifact
is a release-ready 27B model.

\paratitle{Industrial code intelligence.}
We use \emph{industrial coding} to denote code-generation problems whose
acceptance is determined by specialized executable toolchains and deployment
quality, rather than by surface similarity alone
\citep{pinckney2025cvdp,fu2026synthesis,ouyang2025kernelbench,li2025tritonbench,zang2026kernelgenbench,huang2026realistictriton}.
Within hardware design, RTL generation is a code-intensive stage of the broader
EDA workflow: a behavioral specification must become a clocked circuit
description that elaborates, synthesizes, and satisfies its intended behavior
\citep{liu2023verilogeval,lu2023rtllm,pinckney2025cvdp,fu2026synthesis,deng2026acertl}.
Accordingly, evaluation has progressed from short, isolated modules toward
realistic specifications, hierarchical IP designs, and simulation-, formal-,
and synthesis-grounded validation
\citep{liu2023verilogeval,lu2023rtllm,pinckney2025cvdp,jin2025realbench,purini2025archxbench,fu2026synthesis,wang2026rtlbenchmt}.
GPU kernel generation forms a parallel systems-coding regime rather than
another stage of EDA
\citep{ouyang2025kernelbench,li2025tritonbench,jaber2026autokernel,brodsky2026kernelforge}.
Here, a program must compile, preserve numerical
semantics, exploit the target machine efficiently, and deliver measured speedup
\citep{ouyang2025kernelbench,li2025tritonbench,jaber2026autokernel,brodsky2026kernelforge,zang2026kernelgenbench,huang2026realistictriton}.
Recent evaluations have consequently expanded beyond isolated operator
translation to diverse workloads, hardware portability, and integration into
real software frameworks
\citep{ouyang2025kernelbench,li2025tritonbench,jaber2026autokernel,brodsky2026kernelforge,zang2026kernelgenbench,huang2026realistictriton}.
These two regimes test
complementary capabilities: RTL stresses temporal and hierarchical reasoning
under exact behavioral constraints, whereas kernels stress parallelization and
performance reasoning under numerical constraints
\citep{jin2025realbench,purini2025archxbench,fu2026synthesis,ouyang2025kernelbench,zang2026kernelgenbench,huang2026realistictriton}.
Most prior systems study one regime in isolation
\citep{deng2026acertl,jaber2026autokernel,brodsky2026kernelforge}; we use both to assess whether a single model can
support a broader notion of industrial code intelligence.

\section{Human Prior: Executable Research Skills}
\label{sec:human-prior}

\begin{insightbox}{Design Principle: The Bitter Lesson for Model Development}
Sutton's \emph{Bitter Lesson} argues that general methods scale when computation
supports search and learning rather than manually crafted decisions
\citep{sutton2019bitterlesson}. We apply this principle to model development:
humans encode objectives, trusted verifiers, permissions, and reproducibility
as an executable process prior. Within that boundary, the agent itself uses evidence
to choose data, objectives, rewards, and stage transitions.
\end{insightbox}

\subsection{From Expert Experience to an Executable Prior}

We represent the human prior as a versioned collection of \emph{Research
Skills}: operational knowledge that turns a broad model-development objective
into an executable, but deliberately under-specified, research program. The
Skills encode procedures that an experienced team would otherwise communicate
through repeated intervention, including how to access approved resources,
construct auditable training artifacts, launch distributed jobs, diagnose
failures, and consume executable verification. They also define invariants that
must not be learned by trial and error, such as held-out-data separation,
verifier integrity, permission boundaries, and the requirement that every
reported checkpoint be traceable to its data, code, configuration, and parent
model. Table~\ref{tab:human-prior-scope} summarizes this prior.

The prior fixes the objective, stage scaffold, admissible action space, and
run-time bindings for models, repositories, data sources, launchers, verifier
profiles, and resource quotas. The agent selects data transformations,
privileged context, optimization objectives, reward adapters, and checkpoints.
Experimental findings enter the experiment record and inform subsequent
decisions.

\begin{table}[H]
    \centering
    \caption{Scope of the human prior. Research Skills specify reusable
    procedures and non-negotiable boundaries while leaving empirical choices
    to the agent. Run-specific identities and measured findings are recorded
    separately in the project state.}
    \label{tab:human-prior-scope}
    \footnotesize
    \setlength{\tabcolsep}{3.5pt}
    \renewcommand{\arraystretch}{1.12}
    \begin{tabular}{@{}
        >{\raggedright\arraybackslash}p{0.19\linewidth}
        >{\raggedright\arraybackslash}p{0.39\linewidth}
        >{\raggedright\arraybackslash}p{0.34\linewidth}@{}}
        \toprule
        \textbf{Prior layer} & \textbf{Human-specified content}
        & \textbf{Boundary on agent action} \\
        \midrule
        \tableicon{bullseye}Objective and scaffold
        & Target capability, completion criteria, and the Data--SFT--OPSD--RLVR
          stage roles
        & May select experiments and transitions, but may not silently redefine
          the deliverable. \\
        \skillrow{Skill excerpt --- stage controller}{
          data $\longrightarrow$ SFT $\longrightarrow$ OPSD
          $\longrightarrow$ RLVR\\[-0.15em]
          $\hookleftarrow$ returning to data is normal}
        \cdashline{1-3}[1.2pt/1.5pt]
        \tableicon{database}Resource access
        & Approved repositories, models, data, launchers, compute, storage, and
          external-side-effect permissions
        & May use registered capabilities within budget; new identities or
          permissions require a human gate. \\
        \cdashline{1-3}[1.2pt/1.5pt]
        \tableicon{check-circle}Verification
        & Task-contract schema, official-harness requirement, integrity checks,
          and pass/fail/unjudged semantics
        & May integrate and audit a verifier profile; may not weaken or replace
          the approved correctness criterion. \\
        \cdashline{1-3}[1.2pt/1.5pt]
        \tableicon{flask}Research method
        & Controlled pilots, explicit baselines, artifact lineage, stop rules,
          and regression-aware checkpoint selection
        & May formulate hypotheses and choose the next discriminating test;
          conclusions must cite registered evidence. \\
        \cdashline{1-3}[1.2pt/1.5pt]
        \tableicon{clipboard-list}Governance and memory
        & Task Queue, experiment log, decision records, and approval boundaries
        & May write run-scoped findings and negative results; the released
          Human Prior remains immutable during the run. \\
        \skillrow{Condensed executable rule --- decision memory}{
          observe $\rightarrow$ conclude $\rightarrow$ act $\rightarrow$ expect\\[-0.15em]
          return $\rightarrow$ record what actually happened}
        \bottomrule
    \end{tabular}
\end{table}

\subsection{A Staged Prior for Post-Training}

The scaffold separates executable data construction from three parameter-update
stages. Data construction first produces task--oracle pairs whose correctness
can be tested, because every later stage requires a trustworthy signal about
the generated code. SFT then provides a reasoning cold start from verified
teacher trajectories on tasks that expose a model-relative capability gap.
OPSD occupies the intermediate regime in which the current model can solve a
task only after receiving task-specific guidance or feedback. Its intended role
is to convert this observed self-correction into a persistent capability while
keeping the deployed policy conditioned on the original task. RLVR finally
moves supervision online: the policy explores its own responses and receives
task-conditioned rewards derived from executable outcomes. These roles motivate
the order Data $\rightarrow$ SFT $\rightarrow$ OPSD $\rightarrow$ RLVR. The
agent determines the concrete interventions within each stage.

The stage sequence remains revisable. SFT can reveal missing task families and
return the loop to data
construction. OPSD is entered only after evaluation identifies a named,
recoverable weakness and may return upstream when suitable failure experience
cannot be constructed. RLVR requires both a trustworthy verifier and rollout
groups with informative reward variation; otherwise the agent revises the
verification, reward, or data design. A stage transition is accepted only after
the originating checkpoint and its regressions have been evaluated under a
fixed protocol.

\subsection{Shared Resources, Governance, and Memory}

The Skills expose repositories, model and data interfaces, task-native
verifiers, cluster launchers, storage, monitoring, and documentation procedures
through an approved project profile. Resource access is capability-based: the
agent may compose and execute registered components inside the declared compute
and permission envelope. New datasets, teachers, correctness rules, external
side effects, and boundary changes require human authorization. Data policy,
evaluation semantics, and release authority remain human-owned.

Within that envelope, an Agent Task Queue makes the prior actionable. Each item
registers a research question or build objective, its dependencies, executable
action, budget, expected artifacts, acceptance rule, and the decision enabled
by the result. The agent repeatedly observes the current evidence, diagnoses a
limitation, proposes the smallest discriminating experiment, executes it,
verifies its artifacts, and either retains, revises, or rejects the hypothesis.
The resulting decisions and negative results form run-scoped research memory.
This memory changes which task is attempted next and preserves an auditable
boundary between the original human prior and knowledge acquired by the loop.

\section{Agentic Experimentation Trajectory}
\label{sec:agent-loop}

\subsection{Data: Verifier-Guided Evolution of a Shared Executable Task Pool}
\label{sec:agent-loop-data}

\paratitle{Research prior and decision space.}
The Data Research Skill supplied a common executable item contract, a library
of evolution operators, and the admission invariants summarized in
Table~\ref{tab:research-prior-data}. Human experts registered the seed sources
and task-native verifiers. Pilot evidence guided the agent's route selection,
validation-gate order, retry budget, and final pool composition.

\begin{table}[!t]
    \centering
    \caption{Data Research Skill exposed to the agent. The Skill supplies
    executable primitives and admission invariants, while transformation and
    corpus-composition choices remain evidence-driven.}
    \label{tab:research-prior-data}
    \footnotesize
    \setlength{\tabcolsep}{3.4pt}
    \renewcommand{\arraystretch}{1.10}
    \begin{tabular}{@{}
        >{\raggedright\arraybackslash}p{0.19\linewidth}
        >{\raggedright\arraybackslash}p{0.37\linewidth}
        >{\raggedright\arraybackslash}p{0.36\linewidth}@{}}
        \toprule
        \textbf{Skill component} & \textbf{Exposed primitives and invariants}
        & \textbf{Agent decision space} \\
        \midrule
        \tableicon{file-contract}Task contract
        & Specification, generated-artifact boundary, interface, reference
          status, and task-native harness
        & Normalize each seed and determine whether its current oracle is usable. \\
        \skillrow{Condensed executable rule --- item contract}{
          item $:=$ instruction $+$ reference solution $+$ verifier\\[-0.15em]
          require all three parts to agree}
        \cdashline{1-3}[1.2pt/1.5pt]
        \tableicon{random}Evolution operators
        & Behavior-preserving rewriting and reconstruction; contract-bound
          behavioral mutation
        & Select an operator per seed and revise the operator library from pilot failures. \\
        \skillrow{Skill excerpt --- solution-first evolution}{
          mutate reference solution $\rightarrow$ derive instruction\\[-0.15em]
          $\rightarrow$ generate verifier from the same mutation}
        \cdashline{1-3}[1.2pt/1.5pt]
        \tableicon{users-cog}Generation roles
        & Separate proposal, implementation, specification, and audit roles
        & Decide when a teacher-generated candidate can become a verified reference. \\
        \cdashline{1-3}[1.2pt/1.5pt]
        \tableicon{filter}Admission gates
        & Interface preservation, leakage checks, semantic-change tests,
          solvability, compilation, and execution
        & Order cheap and expensive gates, classify failures, and allocate retries. \\
        \skillrow{Condensed executable rule --- rejection pipeline}{
          static gates $\rightarrow$ verifier strength $\rightarrow$ contract audit\\[-0.15em]
          $\rightarrow$ dedup $\rightarrow$ difficulty $\rightarrow$ compose}
        \cdashline{1-3}[1.2pt/1.5pt]
        \tableicon{layer-group}Pool composition
        & Ancestor-level split, lineage, deduplication, difficulty measurement,
          and coverage accounting
        & Set per-lineage caps and balance coverage, difficulty, and cost. \\
        \bottomrule
    \end{tabular}
\end{table}

\begin{figure}[H]
  \centering
  \includegraphics[width=\linewidth]{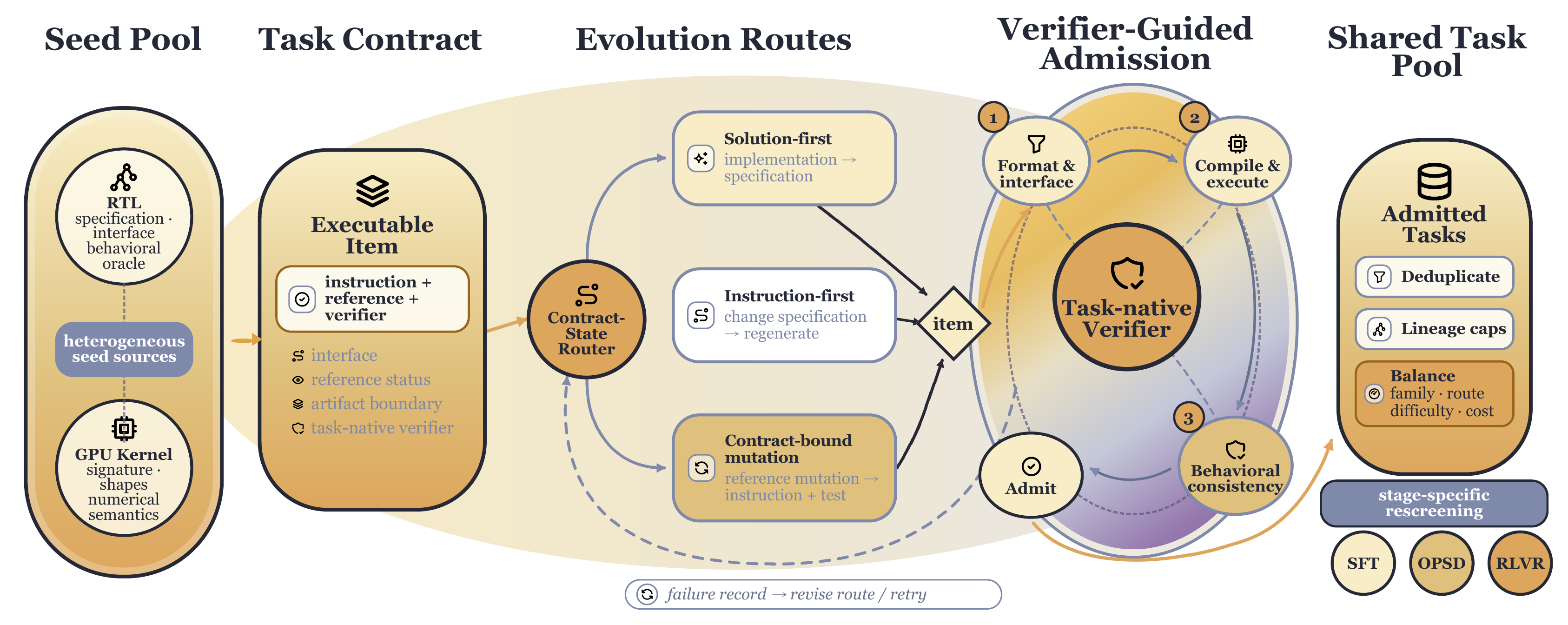}
  \caption{\textbf{Verifier-guided construction of the shared executable task
  pool.} The agent normalizes RTL and GPU-kernel seeds into a common item
  contract, selects among three evolution routes to form candidate task--oracle
  pairs, and applies ordered task-native validation gates. Failures update
  routing and retry decisions; admitted items are deduplicated and balanced
  before stage-specific rescreening for SFT, OPSD, and RLVR.}
  \label{fig:data-collection-pipeline}
\end{figure}
\FloatBarrier

\paratitle{Normalizing heterogeneous seeds.}
The agent represented every seed by its natural-language specification,
generated-artifact boundary, interface, reference status, and trusted verifier.
For RTL tasks, this contract records module ports, clock and reset behavior,
timing requirements, and the behavioral oracle. For GPU kernels, it records the
model interface, argument shapes, numerical semantics, and permitted
implementation boundary. This representation allowed the two domains to share
planning, provenance, and composition logic while retaining task-native
execution.

\paratitle{Routing evolution by contract state.}
The agent routed seeds among three evolution families according to reference
availability and the contract surfaces affected by the transformation.
Solution-first reconstruction derived a new specification from a verified
implementation while preserving its behavior and verifier. Instruction-first
evolution changed the specification and regenerated the affected executable
components before validation. Contract-bound mutation applied a declared
behavioral operator to the reference, then derived a matching instruction and
test from the same change. For the third route, the evolved reference had to
pass the evolved test and the original reference had to fail it. Pilot results
determined scale allocation from verified yield, behavioral distance from the
parent, rejection causes, and cost per surviving item.

\paratitle{Failure-driven domain adaptation.}
RTL pilots showed that compressed specifications often dropped reset, timing,
or state-transition rules. The retained pipeline therefore extracted interfaces
mechanically, preserved explicit clock and reset contracts, and used
structure-aware operators for stateful designs. Kernel pilots exposed a
different pattern: some transformations left the computation unchanged, some
produced ordinary PyTorch instead of Triton, and others drifted in signatures
or tensor shapes. The retained kernel path pinned the model interface and input
contract, checked implementation feasibility, and executed candidates against
the trusted reference. Both paths ordered inexpensive format and interface
checks before compilation and task-native execution. Rejection stages remained
part of the record and guided the next operator, retry policy, and sampling
allocation. Appendix~\ref{app:verifier-design} describes the corresponding
verifier contracts.

\paratitle{Composing and handing off the shared task pool.}
After executable validation, the agent deduplicated candidates and composed the
pool in two passes. It selected one valid candidate per seed lineage, then
filled the remaining budget under per-seed caps while balancing family, route,
difficulty, and verification cost. Each stage rescreened the resulting shared
task pool against its checkpoint and supervision objective;
Appendix~\ref{app:data-analysis} reports its composition.

\subsection{SFT: Cold-Starting Reasoning from Verified Teacher Trajectories}
\label{sec:agent-loop-sft}

\paratitle{Research prior and decision space.}
The SFT Research Skill defined a verified cold-start family. It supplied a
stable prompt--response format,
task-native trajectory verification, loss-mask and length audits, a
production-path smoke test, and composition-aware checkpoint evaluation
(Table~\ref{tab:research-prior-sft}). Within the human-defined capability-gap
family, the agent selected the sampling budget, the student and teacher success
predicates, the retained passing trajectory, the distributed layout, and the
saturation point. These decisions were recorded
against the current model, data pool, and compute profile. The intended outcome
was a minimal reasoning cold start on tasks where the base model showed no
verified success.

\begin{table}[!htbp]
    \centering
    \caption{SFT Research Skill exposed to the agent. The prior fixes data and
    training integrity requirements but leaves the model-relative curriculum,
    scale, and stopping decision to experiment.}
    \label{tab:research-prior-sft}
    \footnotesize
    \setlength{\tabcolsep}{3.4pt}
    \renewcommand{\arraystretch}{1.10}
    \begin{tabular}{@{}
        >{\raggedright\arraybackslash}p{0.19\linewidth}
        >{\raggedright\arraybackslash}p{0.37\linewidth}
        >{\raggedright\arraybackslash}p{0.36\linewidth}@{}}
        \toprule
        \textbf{Skill component} & \textbf{Exposed primitives and invariants}
        & \textbf{Agent decision space} \\
        \midrule
        \tableicon{code}Format contract
        & Prompt template, response boundary, extraction, tokenizer, and loss mask
        & Audit compatibility and reject format changes that invalidate comparison. \\
        \cdashline{1-3}[1.2pt/1.5pt]
        \tableicon{filter}Capability-gap filter
        & Repeated student and teacher sampling with task-native verification
        & Choose sampling budgets and the failure/success predicate defining the subset. \\
        \skillrow{Condensed executable rule --- model-relative filtering}{
          student trials $\leftarrow$ sample(student, task)\\[-0.15em]
          subset $\leftarrow$ select(gap-predicate(student trials))}
        \cdashline{1-3}[1.2pt/1.5pt]
        \tableicon{check-double}Verified target
        & Complete teacher trajectory retained only after an executable pass
        & Select among passing trajectories and record uncovered tasks. \\
        \skillrow{Condensed executable rule --- target admission}{
          candidates $\leftarrow$ sample(teacher, task)\\[-0.15em]
          retain $y$ only if verifier$(y)=$ pass; log uncovered tasks}
        \cdashline{1-3}[1.2pt/1.5pt]
        \tableicon{shield-alt}Training integrity
        & Length audit, no-truncation rule, production-path smoke test,
          checkpoint and telemetry requirements
        & Choose sequence budget, parallel layout, batch configuration, and save cadence. \\
        \skillrow{Condensed executable rule --- length integrity}{
          $B \leftarrow$ empirical length distribution of training pairs\\[-0.15em]
          length $>B$ $\Rightarrow$ drop; never truncate}
        \cdashline{1-3}[1.2pt/1.5pt]
        \tableicon{route}Evaluation and routing
        & Fixed protocol, composition-aware metrics, and failure taxonomy
        & Select the checkpoint, diagnose saturation, and return to Data or enter OPSD. \\
        \bottomrule
    \end{tabular}
\end{table}

\paratitle{Selecting verified teacher trajectories.}
The agent selected tasks with Qwen3.6-27B pass@4 equal to zero and at least one
verified success among four DeepSeek-V4-Pro trajectories. One complete passing
trajectory was retained per task. Full-parameter SFT on this corpus produced
the cold-start reasoning checkpoint.

\FloatBarrier
\subsection{OPSD: Recursive Discovery of Verifier-Anchored Self-Distillation}
\label{sec:agent-loop-opsd}

\paratitle{Research prior and decision space.}
On-policy distillation evaluates teacher supervision on student-generated
trajectories, reducing the train--inference distribution mismatch of off-policy
distillation \citep{agarwal2024gkd}. OPSD specializes this idea to a single
model whose teacher and student views differ only in the context they receive
\citep{zhao2026opsd}. The human-provided Skill exposed this same-weight,
dual-context family as a candidate bridge between supervised cold start and
online RLVR. Runtime errors and judge feedback can condition the current model
into a feedback-informed self-teacher \citep{hubotter2026sdpo}; the Skill
therefore admitted plans, failed attempts, and verifier diagnostics as candidate
evidence, subject to a strict no-solution boundary.

Table~\ref{tab:research-prior-opsd} makes the attribution boundary explicit.
The human prior required a named measured weakness, same-weight teacher and
student views, executable outcome evidence, leakage controls, and instrumentation
that could distinguish a productive update from a merely decreasing loss. The
agent selected the recoverability criterion, the contents and granularity of
the privileged view, the role of contextual preference in update direction and
token credit, and the checkpoint to advance. Controlled ablations and
failure-driven redesign resolved these coupled decisions.

\begin{table}[!htbp]
    \centering
    \caption{OPSD Research Skill exposed to the agent. The prior defines a
    verifier-anchored self-distillation family; data eligibility, privileged
    evidence, and the learning rule are selected through controlled tests.}
    \label{tab:research-prior-opsd}
    \footnotesize
    \setlength{\tabcolsep}{3.4pt}
    \renewcommand{\arraystretch}{1.10}
    \begin{tabular}{@{}
        >{\raggedright\arraybackslash}p{0.19\linewidth}
        >{\raggedright\arraybackslash}p{0.37\linewidth}
        >{\raggedright\arraybackslash}p{0.36\linewidth}@{}}
        \toprule
        \textbf{Skill component} & \textbf{Exposed primitives and invariants}
        & \textbf{Agent decision space} \\
        \midrule
        \tableicon{crosshairs}Target and data
        & Named measured weakness, repeated sampling, execution-feedback loop,
          and recoverability evidence
        & Define the eligibility predicate and rescreen it under the current checkpoint. \\
        \skillrow{Condensed executable rule --- entry contract}{
          require named weakness $+$ baseline measurement\\[-0.15em]
          require teacher advantage relevant to that weakness}
        \cdashline{1-3}[1.2pt/1.5pt]
        \tableicon{eye}Privileged view
        & Plan, failed responses, verifier diagnostics, and matched control contexts
        & Select components, ordering, granularity, and stopping point for $c^\star$. \\
        \cdashline{1-3}[1.2pt/1.5pt]
        \tableicon{lock}Information boundary
        & No reference answer, code-overlap scan, independent audit, and context integrity
        & Reject leaking plans and decide whether a component remains admissible. \\
        \cdashline{1-3}[1.2pt/1.5pt]
        \tableicon{signal}Learning signal
        & Frozen privileged scoring, executable outcome, group baseline, and clipped update family
        & Decide what determines update direction and how teacher evidence assigns token credit. \\
        \skillrow{Skill excerpt --- offline OPSD phases}{
          roll out $\rightarrow$ score $\rightarrow$ weight $\rightarrow$ train\\[-0.15em]
          generation and teacher scoring remain outside the update phase}
        \cdashline{1-3}[1.2pt/1.5pt]
        \tableicon{tachometer-alt}Instrumentation
        & Loss, parameter movement, trajectory behavior, regression matrix, and stop conditions
        & Diagnose ineffective or collapsing objectives and revise or reject the intervention. \\
        \skillrow{Condensed executable rule --- update validity}{
          track(relative parameter update, mean log-ratio, clipped-token fraction)\\[-0.15em]
          assert first-step ratio $\approx 1$; re-roll when drift exceeds tolerance}
        \bottomrule
    \end{tabular}
\end{table}

\paratitle{Selecting a recoverable capability frontier.}
The agent first used repeated sampling to identify tasks with pass@4 equal to
zero under the current SFT checkpoint. It then ran an execution-feedback loop
and retained tasks that failed on the initial bare-prompt attempt but were
solved in a later round. This \emph{round-0-fail and loop-solved} criterion
distinguished recoverable failures from examples that were already mastered and
from failures for which repeated interaction supplied no usable correction.
Applying this criterion produced a pool of 1,874 tasks. These tasks defined a
capability frontier at which the model's own
unsuccessful attempts and verifier feedback were sufficient to reach a
solution.

\paratitle{Constructing privileged context through controlled ablations.}
The agent treated context construction as a design problem with three separate
questions: whether guidance should be task-specific, how much verifier detail
should be exposed, and whether the resulting signal remained useful after
answer leakage was removed. Panel (a) of
Table~\ref{tab:opsd-context-ablation} enumerates every tested context view;
panel (b) reports only matched contrasts. Each contrast uses its own evaluation
set and primary endpoint; cross-row effect comparisons are invalid. Here avg@5 averages five binary
correctness outcomes per task and differs from pass@5; pp denotes percentage
points.

\begin{table}[!t]
    \centering
    \caption{Privileged-context construction and matched ablations. Panel~(a)
    summarizes all context variants considered by the agent and the resulting
    composition of $c^\star$. Panel~(b) reports controlled comparisons for plan
    guidance, verifier diagnostics, feedback granularity, and task--context
    matching. Each comparison uses its own matched evaluation set and should be
    interpreted within the corresponding row.}
    \label{tab:opsd-context-ablation}

    \footnotesize
    \setlength{\tabcolsep}{8.2pt}
    \renewcommand{\arraystretch}{1.13}

    \begin{tabular}{@{}
        >{\raggedright\arraybackslash}p{0.18\linewidth}
        >{\raggedright\arraybackslash}p{0.50\linewidth}
        >{\raggedright\arraybackslash}p{0.14\linewidth}@{}}
        \toprule
        \multicolumn{3}{@{}l}{\textbf{(a) Candidate contexts and final composition}} \\
        \addlinespace[1pt]
        \textbf{Context view} & \textbf{Information beyond the task prompt}
        & \textbf{Disposition} \\
        \midrule
        Bare context $c_{\varnothing}$
        & None; the model receives only the task prompt.
        & Reference \\
        \cdashline{1-3}[1.2pt/1.5pt]

        Direction-only plan $c_{\mathrm{plan}}$
        & A task-specific repair direction; code, assignments, constants, and
          answer-specific mappings are prohibited.
        & Include plan \\
        \cdashline{1-3}[1.2pt/1.5pt]

        Generic domain hint
        & Task-independent domain advice without evidence from the current
          attempt.
        & Exclude \\
        \cdashline{1-3}[1.2pt/1.5pt]

        Error-conditioned experience
        & A failed response paired with its diagnostic verifier output.
        & Include signal \\
        \cdashline{1-3}[1.2pt/1.5pt]

        Coarse experience $c_{\mathrm{coarse}}$
        & Failed responses plus verifier stage or mismatch rate and a general
          repair checklist.
        & Include \\
        \cdashline{1-3}[1.2pt/1.5pt]

        Fine experience $c_{\mathrm{fine}}$
        & $c_{\mathrm{coarse}}$ plus at most three input-level
          expected-versus-observed counterexamples.
        & Exclude detail \\
        \midrule
        Final $c^\star$
        & Audited $c_{\mathrm{plan}}$ plus every failed response and its coarse
          verifier feedback before the first successful round.
        & \textbf{Selected} \\
        \bottomrule
    \end{tabular}

    \vspace{1.4ex}

    \begin{tabular}{@{}
        >{\raggedright\arraybackslash}p{0.13\linewidth}
        >{\raggedright\arraybackslash}p{0.27\linewidth}
        >{\raggedright\arraybackslash}p{0.19\linewidth}
        >{\raggedright\arraybackslash}p{0.23\linewidth}@{}}
        \toprule
        \multicolumn{4}{@{}l}{\textbf{(b) Matched comparisons}} \\
        \addlinespace[1pt]
        \textbf{Question} & \textbf{Treatment / control}
        & \textbf{Evaluation protocol} & \textbf{Primary result} \\
        \midrule
        Plan value
        & $c_{\mathrm{plan}}$ vs. $c_{\varnothing}$
        & Same 194 tasks; five samples per condition
        & avg@5: 47.4\% vs. 41.3\% ($+6.1$ pp) \\
        \cdashline{1-4}[1.2pt/1.5pt]

        Diagnostic value
        & Failure history with diagnostics vs. the same history without them
        & 92 round-0-failure tasks
        & Additional solved tasks: 6/92 ($+6.5$ pp) \\
        \cdashline{1-4}[1.2pt/1.5pt]

        Feedback detail
        & $c_{\mathrm{fine}}$ vs. $c_{\mathrm{coarse}}$
        & Reported 58-task paired contingency
        & Final coverage: 49/58 in both conditions (exact McNemar $p=1.00$) \\
        \cdashline{1-4}[1.2pt/1.5pt]

        Task match
        & Matched $c^\star$ vs. shuffled $c^\star$
        & Within-domain shuffle; same responses rescored
        & Mean response log-likelihood: approximately $+0.18$ nats \\
        \bottomrule
    \end{tabular}

\end{table}

On the same 194 tasks and sampling budget, a task-specific plan improved
avg@5 by 6.1 percentage points; the gain was 8.5 points on round-0
failures and 4.0 points on initially solved tasks. A separate in-distribution
pilot found a net negative effect from task-independent domain hints, which
were therefore excluded from the default context. The feedback experiments
then separated diagnostic value from diagnostic granularity. On the 92-task
round-0-failure pool, removing verifier diagnostics reduced solved-task
coverage by six tasks.
In the granularity comparison, fine and coarse feedback each solved 49 of the
58 tasks represented in the reported paired contingency; the five fine-only
and five coarse-only successes yield an exact McNemar $p$-value of 1.00. Fine
feedback produced four more intermediate loop rescues (23 versus 19) but did
not expand final coverage. The agent therefore selected the smallest context
supported by these interventions:
\begin{equation}
    c_i^\star = c_{\mathrm{plan},i}
    \oplus \bigl\{(y_{i,q}, f_{i,q})\bigr\}_{q < q_i^\star},
    \label{eq:cstar-construction}
\end{equation}
Here $y_{i,q}$ and $f_{i,q}$ are the response and coarse verifier feedback at
failed round $q$, $q_i^\star$ is the first successful round, and $\oplus$
concatenates context in order; the successful response is omitted.

The plan component passed three leakage controls before entering $c^\star$: an
instruction-level prohibition on solution content, a rule-based scan for code
and answer-specific overlap, and an audit by a model different from the plan
generator. A plan that repeatedly failed these checks was removed, leaving only
the audited failure experience. The final no-training signal audit compared
four views of the same response: the bare prompt $c_{\varnothing}$, its matched
$c^\star$, a within-domain shuffled $c^\star$, and a length-matched neutral
context. The matched context exceeded the shuffled control by approximately
0.18 nats. Formatting accounted for only 12--22\% of the largest token-level
differences, while the remaining differences concentrated on code tokens. This
control ruled out context length and formatting as sufficient explanations for
the observed teacher advantage.

\paratitle{Failure of gap-directed optimization.}
The first implementation used the token-level likelihood gap between the
privileged and bare views as the direction of an online policy update. Because
the privileged view was scored by the current actor, its distribution moved
with the student and created a self-reinforcing feedback path. Early
checkpoints improved several shorter-horizon tasks, but the same intervention
regressed harder tasks. Matched ArchXBench trajectories became approximately
2.9 times longer, and some generations repeatedly reproduced an intermediate
reasoning segment without emitting executable code. Continued training
eventually collapsed into degenerate repetition. The agent interpreted this
failure as evidence that contextual preference and environmental correctness
were being conflated. A likelihood difference identified tokens favored under
$c^\star$; executable outcomes determined whether a response should be
reinforced.

\begin{insightbox}{Agent-Discovered Lesson: Privilege Can Allocate Credit, Not Authority}
The privileged view revealed which tokens became more plausible under useful
experience, but not whether the resulting behavior was correct. The failed
online objective exposed a clean separation: executable outcomes must determine
the direction of learning, while privileged context should only redistribute
token-level credit. In self-distillation, information advantage is not
optimization authority.
\end{insightbox}

\paratitle{Separating update direction from token credit.}
The agent autonomously replaced the online loop with an offline five-stage
pipeline: bare-prompt rollout, teacher-forcing scores under $c^\star$,
executable verification, signed-advantage construction, and a clipped
importance-weighted update. Freezing and precomputing the teacher scores broke
the moving-teacher feedback path. For response $i$, the verifier first produced
a raw hierarchical advantage
$A_{i,\mathrm{raw}}^{\mathrm{OPSD}}=
A_i^{\mathrm{valid}}+A_i^{\mathrm{perf}}$.
For kernel tasks, compilation failures received $-1$ and executable but
incorrect responses received $-0.5$; other verifier failures received $-0.5$.
Correct responses received zero validity penalty. Among correct responses, the
performance term used robustly normalized log speedup for kernel tasks and
centered assertion coverage for ArchXBench; it was zero when no graded
performance signal existed. Following the group-relative baseline used by
GRPO \citep{shao2024deepseekmath}, group centering then yielded
$\widetilde A_i^{\mathrm{OPSD}}=A_{i,\mathrm{raw}}^{\mathrm{OPSD}}-
\frac{1}{|G_i|}\sum_{j\in G_i}A_{j,\mathrm{raw}}^{\mathrm{OPSD}}$.
Here $G_i$ is the same-prompt rollout group containing response $i$; candidates
that were relatively better within the group received positive credit.

Let $\ell^T_{it}$ denote the frozen teacher log-likelihood of token $t$ under
$c_i^\star$, $\ell^{\mathrm{old}}_{it}$ the rollout-policy log-likelihood, and
$\ell^\theta_{it}$ the current student log-likelihood; $\operatorname{sg}$
denotes stop-gradient. The three arms supplied different effective advantages
to the same clipped PPO-style update
\citep{schulman2017ppo,shao2024deepseekmath}:
\begin{equation}
\begin{aligned}
    a^{\mathrm{B0}}_{it}
        &= \ell^T_{it}-\ell^{\mathrm{old}}_{it}, \\
    a^{\mathrm{B1}}_{it}
        &= \widetilde A_i^{\mathrm{OPSD}}, \\
    a^{\mathrm{B2}}_{it}
        &= \widetilde A_i^{\mathrm{OPSD}}
           \bigl((1-\lambda)+\lambda w_{it}\bigr), \\
    w_{it}
        &= \operatorname{clip}\!\left(
            \exp\!\left(\operatorname{sign}
            (\widetilde A_i^{\mathrm{OPSD}})
            (\ell^T_{it}-\operatorname{sg}[\ell^\theta_{it}])\right),
            1-\epsilon_w,1+\epsilon_w\right).
\end{aligned}
\label{eq:opsd-three-arms}
\end{equation}
All three arms shared the same implementation atop the verl/HybridFlow
training framework \citep{sheng2024hybridflow}, isolating the learning signal.
B0 asks whether contextual preference can determine the update direction. B1
replaces that direction with executable outcome and assigns uniform token
credit. B2 keeps the B1 direction fixed while using the teacher gap only as a
positive, bounded token-level multiplier. This sign-preserving multiplicative
form adapts RLSD \citep{yang2026rlsd}, which anchors update direction to
verifiable outcomes and uses the positive teacher--student evidence ratio only
to redistribute token-level credit. Comparing B2 with B1 therefore isolates
the incremental contribution of privileged-context credit.

\begin{table}[!t]
    \centering
    \caption{Controlled comparison of OPSD objectives from the same SFT
    checkpoint under matched data and training budget. B0 uses the teacher gap
    as the update direction, B1 uses verifier outcome alone, and B2 augments the
    B1 outcome direction with bounded teacher-based token credit. Higher is
    better. The best result in each row is bold and the second-best distinct
    result is underlined. N/A denotes an unavailable SFT measurement.}
    \label{tab:opsd-objective-ablation}

    \small
    \setlength{\tabcolsep}{5.0pt}
    \renewcommand{\arraystretch}{1.08}
    \begin{tabular}{llcccc}
        \toprule
        \textbf{Benchmark} & \textbf{Metric} & \textbf{SFT} & \textbf{B0} & \textbf{B1} & \textbf{B2} \\
        \midrule
        \multirow{2}{*}{VerilogEval}
        & Spec-to-RTL avg@4 & 73.88 & 57.37 & \underline{79.33} & \textbf{84.70} \\
        & Code-complete avg@4 & \underline{81.73} & 56.09 & 79.97 & \textbf{81.90} \\
        \midrule
        \multirow{2}{*}{RTLLM}
        & Functional avg@4 & \underline{60.50} & 40.50 & 59.00 & \textbf{61.20} \\
        & Syntax avg@4 & \underline{83.50} & 61.50 & 79.50 & \textbf{92.50} \\
        \midrule
        \multirow{2}{*}{ArchXBench}
        & Functional pass@1 & 22.54 & 11.27 & \underline{23.94} & \textbf{36.60} \\
        & Syntax pass@1 & 59.15 & 29.58 & \underline{60.56} & \textbf{67.61} \\
        \midrule
        \multirow{2}{*}{RealBench}
        & Functional pass@1 & \underline{9.00} & 4.67 & 6.00 & \textbf{26.70} \\
        & Syntax pass@1 & 13.67 & 10.67 & \underline{14.00} & \textbf{50.00} \\
        \midrule
        \multirow{3}{*}{KernelBench}
        & L1 correct (\%) & \underline{40} & 29 & \textbf{44} & \underline{40} \\
        & L2 correct (\%) & 53 & 43 & \underline{56} & \textbf{60} \\
        & L3 correct (\%) & 8 & 4 & \underline{9} & \textbf{14} \\
        \bottomrule
    \end{tabular}
\end{table}

The agent selected B2 on an internal validation set assigned by itself before the
benchmark evaluation reported in Table~\ref{tab:opsd-objective-ablation}.
B0 scored below the SFT checkpoint on every metric with an available baseline
and below both outcome-directed arms on every row, rejecting gap-directed
optimization. Relative to B1, B2 improved 10 of the 11 reported metrics: both
VerilogEval, RTLLM, ArchXBench, and RealBench metrics, together with KernelBench
L2--L3. The largest gains appeared on RealBench syntax and functional
correctness, which increased by 36.0 and 20.7 points, respectively; ArchXBench
functional pass@1 rose by 12.66 points. KernelBench L1 was the sole regression,
falling from 44\% to 40\%. These results corroborate the validation-based
selection of B2 and demonstrate that bounded teacher-based token credit
transfers across RTL and kernel objectives.

\begin{insightbox}{Research Bet: OPSD as the Sweet Spot Between SFT and RL}
OPSD has drawn broad skepticism in the community because its gains can be unstable, yet our team
maintained a long-standing research bet: with well-designed privileged context,
OPSD can occupy a practical sweet spot between SFT and RL, rapidly repairing
capability gaps and laying a stronger foundation for subsequent RL optimization.
This role has rarely entered the core training pipelines of frontier models.
In this experiment, this judgment was supplied to the agent as a research prior, and the agent-led
RSI loop turned our bet into an effective training recipe, yielding satisfactory performance gains and
practical solutions that we believe can inform future model training.
\end{insightbox}

\FloatBarrier
\subsection{RLVR: Recursive Refinement under Execution-Grounded Rewards}
\label{sec:agent-loop-rlvr}

\paratitle{Research prior and decision space.}
The RLVR stage initialized the policy from the OPSD checkpoint and optimized
responses against rewards obtained by compiling and executing generated code.
The human-provided Skill specified the executable-reward family, task-property
verifier contracts, an immutable raw pass/fail/unjudged verdict, the available
training substrate, and the compute envelope. It required judged correctness to
occupy a separate state from infrastructure failure and preserved the
underlying evidence behind each stage-specific reward scalar.

Within these invariants, the agent controlled the decisions summarized in
Table~\ref{tab:research-prior-rlvr}: which behaviors constituted reward
exploitation for each task family, where hard eligibility gates applied,
whether the available evidence supported binary or graded reward, which failure
classes were retryable or unjudgeable, how missing supervision entered the
group-relative estimator, and how trajectory length and task mixture were
budgeted. Successive pilots and training runs revised both reward validity and
optimization.

\begin{table}[!htbp]
    \centering
    \caption{RLVR Research Skill exposed to the agent. Raw executable verdicts
    remain stable, while eligibility, reward adaptation, and optimization are
    revised only through registered experiments.}
    \label{tab:research-prior-rlvr}
    \footnotesize
    \setlength{\tabcolsep}{3.4pt}
    \renewcommand{\arraystretch}{1.10}
    \begin{tabular}{@{}
        >{\raggedright\arraybackslash}p{0.19\linewidth}
        >{\raggedright\arraybackslash}p{0.37\linewidth}
        >{\raggedright\arraybackslash}p{0.36\linewidth}@{}}
        \toprule
        \textbf{Skill component} & \textbf{Exposed primitives and invariants}
        & \textbf{Agent decision space} \\
        \midrule
        \tableicon{gavel}Executable verdict
        & Task-property contract, official harness profile, positive-pass
          evidence, and pass/fail/unjudged record
        & Integrate profiles and decide when integrity is sufficient for online use. \\
        \cdashline{1-3}[1.2pt/1.5pt]
        \tableicon{shield-alt}Reward eligibility
        & Candidate-independent failure attribution and task-scoped exploit detectors
        & Identify reward exploits and determine where each hard gate applies. \\
        \skillrow{Condensed executable rule --- exploit gate}{
          if exploit detected: reward $\leftarrow 0$\\[-0.15em]
          else: reward $\leftarrow$ task adapter(raw verdict)}
        \cdashline{1-3}[1.2pt/1.5pt]
        \tableicon{sliders-h}Reward adapter
        & Versioned mapping from judged evidence to binary or graded scalar rewards
        & Decide whether meaningful partial or performance ordering exists for a task family. \\
        \cdashline{1-3}[1.2pt/1.5pt]
        \tableicon{question-circle}Missing supervision
        & Closed infrastructure taxonomy, bounded retry, sentinel, and masking contract
        & Classify failures, set stop thresholds, and prevent system faults from becoming gradients. \\
        \skillrow{Condensed executable rule --- unjudged contract}{
          unjudged $\rightarrow$ sentinel outside the legitimate reward range\\[-0.15em]
          exclude from group statistics; set $A=0$; normalize by valid count}
        \cdashline{1-3}[1.2pt/1.5pt]
        \tableicon{cogs}Learning and resources
        & Per-task success rates, four policy-relative strata, reward-variance
          diagnostics, trajectory-length telemetry, and production smoke tests
        & Refresh strata, schedule sampling across difficulty levels, and select
          the estimator, joint trajectory budget, and checkpoint. \\
        \skillrow{Condensed executable rule --- resource and routing checks}{
          budget $\leftarrow |\text{prompt}|+|\text{response}|$\\[-0.15em]
          policy refresh $\Rightarrow$ re-estimate verified success $\Rightarrow$ update sampler;
          excessive unjudged rate $\Rightarrow$ halt}
        \bottomrule
    \end{tabular}
\end{table}

\paratitle{Constructing a policy-aligned task curriculum.}
The agent first ran the OPSD checkpoint over the shared executable task pool
and estimated each task's success rate as the fraction of rollouts receiving
its verified success reward. It then partitioned the pool into four
policy-relative strata: frontier
($0$--$25\%$ success), exploration ($25$--$50\%$), near-mastery
($50$--$75\%$), and exploitation ($75$--$100\%$). Early training emphasized
near-mastery and exploitation tasks to maintain a reliable reward signal. As
success rates increased, the sampler shifted toward exploration and frontier
tasks. During RLVR, the latest policy periodically repeated this evaluation;
the agent refreshed the strata and adjusted the sampling distribution. This
automatic curriculum aligned the training data with the current policy and
kept RLVR near the moving capability boundary of real-world executable tasks
\citep{zhou2025maiui}.

\subsubsection{Kernel Domain: Task-Conditioned Reward Eligibility}
\label{sec:rlvr-kernel-exploration}

\paratitle{Failure of naive execution-based reward.}
The initial kernel pilot accepted numerical correctness together with the
presence of a Triton kernel. The policy found cheaper paths to reward: some
responses delegated computation to the reference framework behind a nominal
Triton wrapper, while others launched an identity kernel that copied loaded
values to the output. Both reproduced the expected output without performing
the requested computation. A graded penalty would retain a positive direction
toward these exploits. The loop therefore made task-specific eligibility a
precondition for assigning reward from the executable outcome.

\paratitle{Scoping eligibility to the task definition.}
Runtime launch evidence, identity-computation detection, and
framework-delegation detection became hard eligibility gates. A detected
exploit received zero reward regardless of numerical output. Uniform scoping,
however, failed. Triton-only tasks require computation inside generated kernels,
whereas KernelBench permits partial operator replacement and evaluates
correctness and speedup. A global delegation gate consequently rejected valid
KernelBench solutions with genuine speedups. The retained rule restricts the
delegation gate to task definitions that prohibit delegation.

Judged and eligible kernel trajectories received the following reward:
\begin{equation}
r_i^{\mathrm{K}} =
\begin{cases}
1.0, & \text{correct with speedup}>1.05\times,\\
0.5, & \text{correct},\\
0.2, & \text{compiled but incorrect},\\
0.05, & \text{failed to compile},\\
0, & \text{failed an exploit-eligibility gate}.
\end{cases}
\label{eq:rlvr-kernel-reward}
\end{equation}
Intermediate reward levels represented policy progress. The eligibility gate
prevented a successful exploit from outranking a genuine implementation.

\subsubsection{RTL Domain: Fail-Closed Behavioral Supervision}
\label{sec:rlvr-rtl-exploration}

\paratitle{Replacing graded progress with behavioral validity.}
RTL tasks exposed a different reward-design problem. Their official harnesses
define success through compilation, simulation, and task-specific functional
checks, but provide no calibrated analogue of kernel speedup. Compilation alone
provides no evidence of partial progress toward behavioral equivalence. The loop
therefore retained a binary outcome for every judged RTL trajectory and reserved
an out-of-range value for a missing verdict:
\begin{equation}
r_i^{\mathrm{RTL}} =
\begin{cases}
1, & \text{the official harness explicitly validates behavior},\\
0, & \text{the candidate is judged and fails},\\
-1, & \text{no verdict remains after an infrastructure retry}.
\end{cases}
\label{eq:rlvr-rtl-reward}
\end{equation}
\paratitle{Diagnosing false verdicts in RTL execution.}
Integration of the RTL backends revealed two ways in which a simulator outcome
could be mistaken for supervision. First, a self-checking testbench could exit
without producing its explicit positive verdict. Treating the absence of a
failure message as success rewarded executions that terminated before
completing their checks. The retained rule is fail-closed: unless an event is
classified as an infrastructure failure, only an explicit positive verdict
receives reward one. Second, malformed payloads and mismatched toolchains could
fail before the generated design was evaluated. In one ArchXBench integration,
the testbench attempted to open a missing input before reaching the candidate;
in CVDP, simulator-version mismatch changed the supported SystemVerilog
constructs. These events carried no evidence about policy quality and entered
the unjudgeable path. Appendix~\ref{app:evaluation-infrastructure}
reports the harness, payload, and toolchain controls that implement this
distinction.

\subsubsection{Cross-Domain Learning: Masked GRPO and Trajectory Budgeting}
\label{sec:rlvr-cross-domain-exploration}

\paratitle{Separating policy failure from infrastructure failure.}
The distinction exposed most clearly by RTL applies to both domains. Compiler
crashes, simulation timeouts, GPU failures, dropped service requests, and
malformed payloads can all prevent a verdict. Scoring every such event as zero
systematically penalizes the affected task family and makes a broken backend
indistinguishable from a persistent model capability gap. The loop introduced
a closed-set infrastructure-error taxonomy: only explicitly enumerated system
failures entered a retry path, while other execution failures remained
attributable to the generated response. A trajectory that could not be judged
after retry received the sentinel $r_i=-1$, outside the legitimate reward interval
$[0,1]$ and denotes missing supervision.

\paratitle{Masking unjudgeable trajectories in group-relative learning.}
The sentinel required a corresponding change inside the trainer. Replacing an
unjudgeable reward with the group mean would preserve the mean but reduce the
estimated variance, thereby inflating the normalized advantages of valid
siblings. Retaining $-1$ as an ordinary reward would create a spurious negative
gradient. The agent excluded unjudgeable samples from both group statistics and
loss normalization. For rollout group $G$, let $r_i$ denote the applicable
kernel or RTL reward, $m_i=\mathbf{1}[r_i\in[0,1]]$, and
$N_G=\sum_{i\in G}m_i$. When $N_G>1$, the masked statistics and advantages,
with stabilizer $\epsilon_{\mathrm{std}}>0$, are
\begin{equation}
\begin{aligned}
\mu_G
    &= \frac{1}{N_G}\sum_{i\in G}m_i r_i,\\
\sigma_G^2
    &= \frac{1}{N_G-1}\sum_{i\in G}m_i(r_i-\mu_G)^2,\\
A_i^{\mathrm{RLVR}}
    &= m_i\frac{r_i-\mu_G}{\sigma_G+\epsilon_{\mathrm{std}}}.
\end{aligned}
\label{eq:rlvr-masked-advantage}
\end{equation}
The policy loss was divided by the number of valid trajectories. Groups with
$N_G\leq1$ produced zero advantages. Because each rollout group contains responses to the same
prompt, the graded kernel reward and binary RTL reward define within-task
orderings; their absolute values are never compared across domains. The
resulting estimator applies the GRPO group-relative update to trajectories with
executable supervision \citep{shao2024deepseekmath}.

Table~\ref{tab:rlvr-decision-trajectory} distinguishes domain-specific reward
validity from shared optimization decisions.
\begin{table}[t]
    \centering
    \caption{Failure-driven refinement of RLVR across kernel and RTL tasks.
    Each row links an observation available to the agent to its diagnosis and
    the intervention retained for the final training run. Absolute rewards are
    compared only within rollout groups for the same task. Verifier
    implementation details are reported separately in
    Appendix~\ref{app:evaluation-infrastructure}.}
    \label{tab:rlvr-decision-trajectory}

    \footnotesize
    \setlength{\tabcolsep}{3.2pt}
    \renewcommand{\arraystretch}{1.08}
    \begin{tabular}{@{}
        >{\raggedright\arraybackslash}p{0.20\linewidth}
        >{\raggedright\arraybackslash}p{0.27\linewidth}
        >{\raggedright\arraybackslash}p{0.25\linewidth}
        >{\raggedright\arraybackslash}p{0.19\linewidth}@{}}
        \toprule
        \textbf{Observed signal} & \textbf{Diagnosis}
        & \textbf{Retained intervention} & \textbf{Role in RLVR} \\
        \midrule
        \multicolumn{4}{@{}>{\columncolor{accent!8}}l@{}}{\strut\textbf{\textcolor{accent}{Kernel-specific reward validity}}} \\
        \cmidrule(lr){1-4}
        Correct output without genuine kernel computation
        & Surface correctness could be achieved without performing the specified task.
        & Converted launch, identity, and scoped delegation checks into hard eligibility gates.
        & Prevent reward exploitation. \\
        \cdashline{1-4}[1.2pt/1.5pt]

        Genuine KernelBench speedups rejected
        & A uniform delegation rule contradicted the benchmark's task definition.
        & Restricted delegation gating to backends that prohibit framework computation.
        & Preserve valid partial replacements. \\
        \midrule

        \multicolumn{4}{@{}>{\columncolor{accent!8}}l@{}}{\strut\textbf{\textcolor{accent}{RTL-specific behavioral validity}}} \\
        \cmidrule(lr){1-4}
        Testbench exited without an explicit positive verdict
        & Absence of a failure message was being interpreted as functional correctness.
        & Required an explicit pass signal and otherwise failed closed.
        & Prevent false-positive RTL rewards. \\
        \cdashline{1-4}[1.2pt/1.5pt]

        Harness failed before evaluating the candidate
        & Missing payloads or mismatched toolchains contained no evidence about policy quality.
        & Routed enumerated pre-candidate failures to retry and the unjudgeable sentinel.
        & Prevent verifier faults from becoming RTL supervision. \\
        \midrule

        \multicolumn{4}{@{}>{\columncolor{accent!8}}l@{}}{\strut\textbf{\textcolor{accent}{Cross-domain optimization}}} \\
        \cmidrule(lr){1-4}

        Persistent zero rewards caused by system faults
        & Infrastructure failure was being learned as policy failure.
        & Added retry, an out-of-range sentinel, and masked group statistics and loss.
        & Remove spurious gradients. \\
        \cdashline{1-4}[1.2pt/1.5pt]

        Most RealBench validation tasks disappeared under the prompt cap
        & Length filtering changed the evaluated RTL distribution without invalidating the run.
        & Raised prompt coverage from the measured distribution before setting the joint budget.
        & Preserve the intended task distribution. \\
        \cdashline{1-4}[1.2pt/1.5pt]

        Long responses approached zero reward and triggered actor OOM
        & Repetition consumed generation time, while per-sequence length set peak memory.
        & Bounded prompt and response jointly for every trajectory.
        & Remove an unproductive tail and stabilize training. \\
        \bottomrule
    \end{tabular}
\end{table}

\paratitle{Diagnosing the trajectory-length failure mode.}
The next experiments showed that context length affected both the task
distribution and training stability. At the initial prompt budget, the training
filter removed only a small number of examples but silently discarded more than
two thirds of the RealBench validation split. The reported score therefore
described only the shortest residual RTL tasks. Raising
the prompt allowance recovered the missing tasks. Independently increasing the
response cap yielded no additional useful training signal. Across tens
of thousands of rollouts, mean reward decreased monotonically from approximately
$0.27$ in the shortest length bucket to about $0.03$ in the middle and $0.02$
near the cap. The longest tail contained no full-reward responses and repeated
intermediate reasoning or code fragments.

The same tail caused an actor-backward out-of-memory failure. The micro-batch
already contained one sequence per GPU, so reducing batch size could not remove
a peak created by a single trajectory. Prompt and response caps were controlling
one physical object through independent limits. The agent replaced them with a
whole-trajectory constraint
\begin{equation}
    \lvert x_i\rvert + \lvert y_i\rvert \leq B_{\mathrm{traj}},
    \label{eq:rlvr-trajectory-budget}
\end{equation}
where $x_i$ and $y_i$ denote the prompt and response. This allocation admits
long RTL prompts without silently changing their evaluation distribution, gives
short prompts more response capacity when useful, and preserves a per-trajectory
memory bound for both domains. It also removes generation after empirical reward
has approached zero.

\paratitle{Closing the recursive RLVR loop.}
Together, these failure-driven revisions formed the RLVR loop.

\begin{insightbox}{Research Outlook: Internalizing Research Taste}
The agent taught us that research skills and research ability are
different layers. Research Skills made hypotheses testable, reproducible, and
safe, but the strongest starting bets still came from humans; once a promising
axis was exposed, the agent was remarkably effective at following failures into
new designs. The next frontier is to internalize research taste itself into the
base model, so agents can not only explore a human-specified research program,
but also invent the hypotheses that expand it.
\end{insightbox}

\section{Methodology}
\label{sec:methodology}

The preceding section records the agent's full development trajectory,
including alternative hypotheses, failed trials, and evidence-driven revisions
across data, SFT, OPSD, and RLVR. This section distills that trajectory into the
retained recipe and presents the end-to-end training pipeline that ultimately
delivered the \ourmethod checkpoint.

\begin{figure*}[!t]
    \centering
    \includegraphics[width=\textwidth]{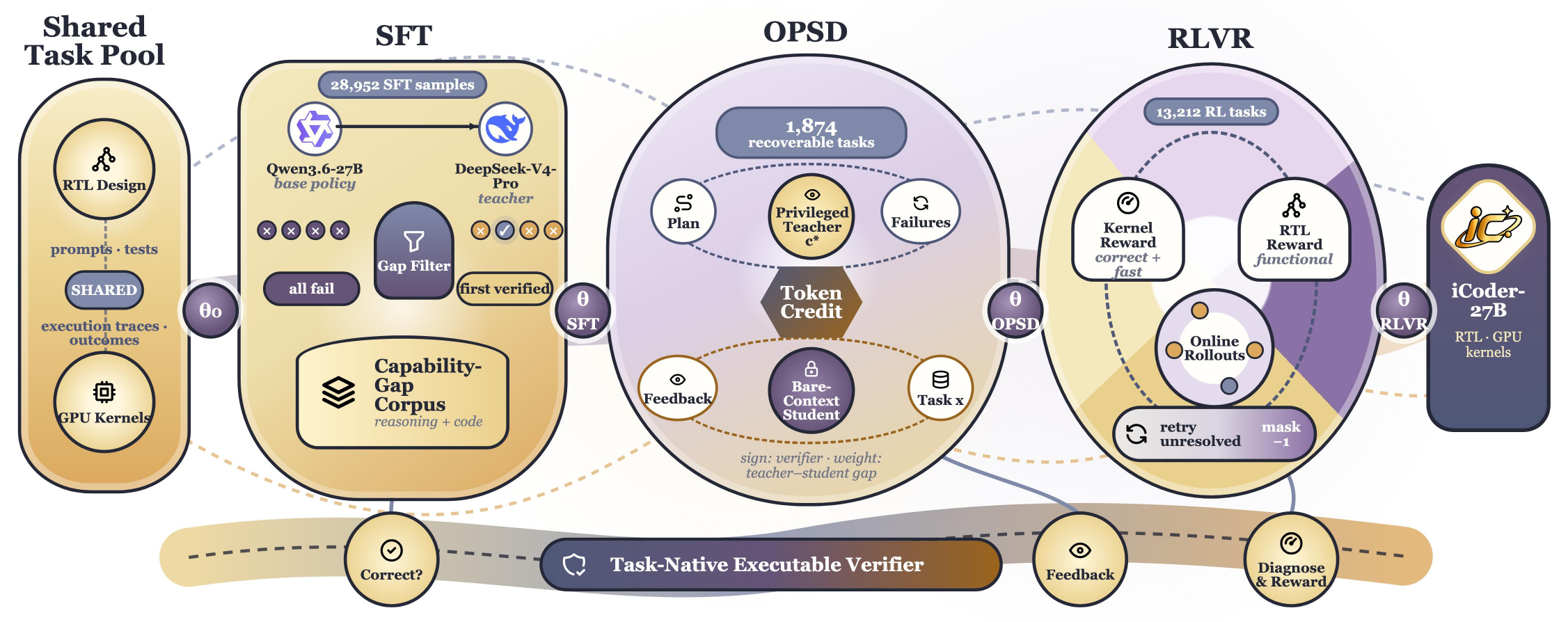}
    \caption{\textbf{Overview of training pipeline.} The shared executable task pool supports three successive parameter-update stages. SFT builds a capability-gap corpus from the first verified teacher trajectory; OPSD transfers privileged experimental context to a bare-context student through verifier-signed token credit; and RLVR converts online execution outcomes into domain-aware rewards while masking unresolved trajectories. The task-native verifier changes roles across stages, from correctness checking, to feedback construction, to failure diagnosis and reward assignment, yielding iCoder-27B.}
    \label{fig:methodology-pipeline}
\end{figure*}

Our final recipe operates on the shared executable task pool constructed from
RTL-design and GPU-kernel tasks. A task-native verifier couples three successive
parameter-update stages:
\[
\theta_0 \xrightarrow{\mathrm{SFT}} \theta_{\mathrm{SFT}}
\xrightarrow{\mathrm{OPSD}} \theta_{\mathrm{OPSD}}
\xrightarrow{\mathrm{RLVR}} \theta_{\mathrm{RLVR}}.
\]
During SFT, verifier decisions
define corpus membership. For each task $x_i$, we draw four responses
$y^B_{i,k}$ from the Qwen3.6-27B base model. Tasks with four failed samples
proceed to teacher rollout, where DeepSeek-V4-Pro produces four trajectories
$y^T_{i,k}$. The first verified teacher trajectory is retained in full,
including its reasoning and executable answer. Writing
$v_i(y)\in\{0,1\}$ for the task-native verdict, the capability-gap corpus and
full-parameter SFT objective are
\begin{equation}
\begin{aligned}
\mathcal I_{\mathrm{SFT}}
    &= \left\{i:\sum_{k=1}^{4}v_i(y^B_{i,k})=0,\;
       \sum_{k=1}^{4}v_i(y^T_{i,k})\geq 1\right\},\\
k_i^\star
    &= \min\{k:v_i(y^T_{i,k})=1\},\qquad
\mathcal D_{\mathrm{SFT}}
    = \{(x_i,y^T_{i,k_i^\star}):i\in\mathcal I_{\mathrm{SFT}}\},\\
\theta_{\mathrm{SFT}}
    &= \arg\min_{\theta}\;-
       \mathbb E_{(x,y)\sim\mathcal D_{\mathrm{SFT}}}
       \sum_{t=1}^{|y|}\log\pi_\theta(y_t\mid x,y_{<t}),
       \qquad \theta\ \text{initialized from}\ \theta_0.
\end{aligned}
\label{eq:method-sft}
\end{equation}
This model-relative filter concentrates cold-start training on base-model
pass@4 failures paired with a verified teacher trajectory
\citep{qwen2026qwen36,deepseek2026v4}.

OPSD rescreens under $\theta_{\mathrm{SFT}}$ and retains 1,874 tasks that are
unsolved at pass@4, fail the first bare-prompt attempt, and recover in a later
execution-feedback round. For each task, $c_i^\star$ comprises the audited
direction-only plan followed by every failed response and its coarse verifier
feedback, terminating before the successful answer
(Eq.~\eqref{eq:cstar-construction}). A frozen copy of
$\theta_{\mathrm{SFT}}$ scores each bare-prompt rollout under
$(x_i,c_i^\star)$. The trainable student conditions on $x_i$. Let
$\ell^T_{it}$, $\ell^{\mathrm{old}}_{it}$, and
$\ell^\theta_{it}$ be the token log-likelihoods of the frozen privileged
teacher, bare-context rollout policy, and current bare-context student, and let
$\widetilde A_i^{\mathrm{OPSD}}$ be the group-centered executable-outcome
advantage.
\begin{equation}
\begin{aligned}
w_{it}
    &= \operatorname{clip}\!\left(
       \exp\!\left[\operatorname{sign}(\widetilde A_i^{\mathrm{OPSD}})
       \bigl(\ell^T_{it}-\operatorname{sg}[\ell^\theta_{it}]\bigr)\right],
       1-\epsilon_w,1+\epsilon_w\right),\\
a^{\mathrm{B2}}_{it}
    &= \widetilde A_i^{\mathrm{OPSD}}
       \bigl((1-\lambda)+\lambda w_{it}\bigr),\qquad
\rho_{it}(\theta)
    = \exp\!\left(\ell^\theta_{it}-\ell^{\mathrm{old}}_{it}\right),\\
\mathcal L_{\mathrm{OPSD}}(\theta)
    &= -\mathbb E_{i,t}\!\left[
       \min\!\left(
       \rho_{it}a^{\mathrm{B2}}_{it},
       \operatorname{clip}(\rho_{it},1-\epsilon_\pi,1+\epsilon_\pi)
       a^{\mathrm{B2}}_{it}\right)\right].
\end{aligned}
\label{eq:method-opsd-b2}
\end{equation}
Equation~\eqref{eq:method-opsd-b2} gives the selected B2 update, where
$\operatorname{sg}$ denotes stop-gradient. The verifier advantage sets the
update sign, and the teacher--student gap defines a positive clipped weight
that redistributes token-level credit across the trajectory
\citep{schulman2017ppo,shao2024deepseekmath,yang2026rlsd}.

RLVR continues from $\theta_{\mathrm{OPSD}}$ and converts executable evidence
into online, task-conditioned rewards. Eligible kernel responses receive the
tiered correctness-and-performance reward in
Eq.~\eqref{eq:rlvr-kernel-reward}. RTL responses receive the binary behavioral
reward in Eq.~\eqref{eq:rlvr-rtl-reward}. Group-relative advantages are
computed within each task's rollout group. Trajectories with unresolved
verification after retry receive the sentinel $-1$. The masking rule in
Eq.~\eqref{eq:rlvr-masked-advantage} removes them from the group mean, standard
deviation, gradient, and loss normalization. Let $\mathcal V$ denote the
verified trajectories retained by masking and $A_i^{\mathrm{RLVR}}$ the
group-relative advantages computed over $\mathcal V$. RLVR uses the GSPO
sequence ratio and asymmetrically clipped objective \citep{zheng2025gspo}:
\begin{equation}
\begin{aligned}
s_i(\theta)
    &= \exp\!\left[
       \frac{1}{|y_i|}\sum_{t=1}^{|y_i|}
       \log\frac{\pi_\theta(y_{i,t}\mid x_i,y_{i,<t})}
       {\pi_{\theta_{\mathrm{old}}}(y_{i,t}\mid x_i,y_{i,<t})}
       \right],\\
\mathcal L_{\mathrm{GSPO}}(\theta)
    &= -\frac{1}{|\mathcal V|}\sum_{i\in\mathcal V}
       \min\!\left(
       s_i(\theta)A_i^{\mathrm{RLVR}},
       \operatorname{clip}\!\left(
       s_i(\theta),1-\epsilon_{\mathrm{lo}},1+\epsilon_{\mathrm{hi}}
       \right)A_i^{\mathrm{RLVR}}\right).
\end{aligned}
\label{eq:method-gspo}
\end{equation}
The implementation broadcasts each sequence advantage across its response
tokens and applies sequence-mean/token-mean aggregation. Entropy weight is
zero, and KL regularization is disabled. All-invalid groups contribute zero
gradient. Every
rollout obeys the whole-trajectory budget in
Eq.~\eqref{eq:rlvr-trajectory-budget}. RLVR yields the final \ourmethod
checkpoint $\theta_{\mathrm{RLVR}}$.

\section{Evaluation}
\label{sec:results}

\subsection{Experimental Setup}
\label{sec:experimental-setup}

\paratitle{Model and post-training data.}
We use Qwen3.6-27B as the base model \citep{qwen2026qwen36}.
Table~\ref{tab:posttraining-data} summarizes the training data used at each
stage.

\begin{table}[h]
    \centering
    \caption{Post-training data by stage. SFT counts complete verified teacher
    trajectories, OPSD counts recoverable executable tasks, and RLVR counts
    stratified verified tasks.}
    \label{tab:posttraining-data}
    \footnotesize
    \setlength{\tabcolsep}{4pt}
    \renewcommand{\arraystretch}{1.15}
    \begin{tabular}{@{}l
        >{\raggedright\arraybackslash}p{5.0cm}
        r r r@{}}
        \toprule
        \textbf{Stage} & \textbf{Training unit}
        & \textbf{Total} & \textbf{RTL} & \textbf{Kernel} \\
        \midrule
        SFT
        & Verified teacher trajectory
        & 28,952
        & 18,256 (63.0\%)
        & 10,696 (37.0\%) \\

        OPSD
        & Recoverable executable task
        & 1,874
        & 1,349 (72.0\%)
        & 525 (28.0\%) \\

        RLVR
        & Stratified verified task
        & 13,212
        & 6,058 (45.8\%)
        & 7,154 (54.2\%) \\
        \bottomrule
    \end{tabular}
\end{table}

\paratitle{Benchmarks and metrics.}
We evaluate RTL generation on VerilogEval, RTLLM, CVDP, ArchXBench, and
RealBench-Module
\citep{liu2023verilogeval,lu2023rtllm,pinckney2025cvdp,purini2025archxbench,jin2025realbench},
and GPU-kernel generation on KernelBench and TritonBench-G
\citep{ouyang2025kernelbench,li2025tritonbench}. The RTL suite ranges from
single-module completion and generation to architecture-level and
project-integrated module tasks; the kernel suite spans single operators,
operator fusion, network-level workloads, and repository-derived Triton
kernels. We follow each benchmark's official task definition and executable
harness, reporting avg@$k$ or pass@$k$ for RTL, compiled/correct/fast rates for
KernelBench, and correctness pass@1 for TritonBench-G. Task counts, sampling
budgets, and verifier settings are provided in
Appendix~\ref{app:evaluation-infrastructure}
(Table~\ref{tab:evaluation-infrastructure}).

\subsection{Main Results}

\begin{table}[!t]
    \centering
    \caption{Performance comparison on industrial coding benchmarks. The best result in each row is shown in bold, and the second-best distinct result is underlined; ties receive the same formatting.}
    \label{tab:industrial-coding-results}

    \setlength{\tabcolsep}{2.4pt}
    \renewcommand{\arraystretch}{1.08}

    \resizebox{\linewidth}{!}{
    \begin{tabular}{lll*{11}{c}}
        \toprule

        \textbf{Benchmark}
        & \multicolumn{2}{c}{\textbf{Metric}}
        & \makecell{\textbf{\ourmethod}\\\textbf{27B}}
        & \makecell{\textbf{Qwen3.6}\\\textbf{27B}}
        & \makecell{\textbf{InCoder}\\\textbf{32B}}
        & \makecell{\textbf{InCoder-32B}\\\textbf{Thinking}}
        & \makecell{\textbf{DeepSeek}\\\textbf{V4-Pro}}
        & \makecell{\textbf{GLM}\\\textbf{5.2}}
        & \makecell{\textbf{Kimi}\\\textbf{K2.6}}
        & \makecell{\textbf{GPT}\\\textbf{5.5}}
        & \makecell{\textbf{Claude}\\\textbf{Opus-4.8}}
        & \makecell{\textbf{Hy3}\\\strut}
        & \makecell{\textbf{Gemini}\\\textbf{3.5-Flash}}
        \\

        \midrule

        \multirow{2}{*}{VerilogEval}
        & \multicolumn{2}{l}{Spec-to-RTL avg@4}
        & 86.3 & 70.1 & 62.5 & 65.9 & 69.9 & 66.0 & 72.4 & \textbf{90.1} & 82.7 & 83.8 & \underline{89.1}
        \\

        & \multicolumn{2}{l}{Code-complete avg@4}
        & \underline{86.0} & 70.8 & 58.2 & 54.2 & 79.8 & 74.8 & 78.5 & \textbf{91.4} & 81.9 & 81.6 & 83.8
        \\

        \midrule

        RTLLM
        & \multicolumn{2}{l}{Functional avg@4}
        & \textbf{68.0} & 49.6 & 48.0 & 44.2 & \underline{67.5} & 64.0 & 59.0 & 66.0 & 64.7 & 53.5 & 63.5
        \\

        \midrule

        CVDP
        & \multicolumn{2}{l}{Functional avg@5 (\%)}
        & \underline{44.1} & 33.9 & 36.9 & 30.3 & 38.5 & 39.5 & 42.1 & 39.5 & \textbf{47.7} & 39.7 & 29.7
        \\

        \midrule

        \multirow{2}{*}{RealBench}
        & \multicolumn{2}{l}{Syntax pass@5 (\%)}
        & 61.7 & 38.3 & 60.0 & 55.0 & 36.7 & 43.3 & 58.3 & \underline{80.0} & \textbf{83.3} & 41.7 & 68.3
        \\

        & \multicolumn{2}{l}{Functional pass@5 (\%)}
        & 26.7 & 16.7 & \textbf{46.7} & \underline{36.7} & 16.7 & 25.0 & 25.0 & 28.3 & \underline{36.7} & 16.7 & 26.7
        \\

        \midrule

        ArchXBench
        & \multicolumn{2}{l}{Functional pass@1 (\%)}
        & 49.3 & 35.2 & 36.6 & 29.6 & 50.7 & 50.7 & 42.3 & \textbf{56.3} & \underline{54.9} & 47.9 & 50.7
        \\

        \midrule

        \multirow{9}{*}{KernelBench}
        & \multirow{3}{*}{L1}
        & Compiled (\%)
        & 95 & 87 & 88 & 85 & 93 & \underline{96} & 93 & \textbf{98} & 95 & 94 & 94
        \\

        & & Correct (\%)
        & \textbf{61} & 32 & 51 & 47 & 32 & 50 & 32 & 43 & \underline{55} & 42 & 45
        \\

        & & Fast (\%)
        & 25 & 12 & 18 & 18 & 13 & \underline{26} & 5 & 22 & \textbf{30} & 21 & 23
        \\

        \cmidrule(lr){2-14}

        & \multirow{3}{*}{L2}
        & Compiled (\%)
        & 97 & 89 & 90 & 93 & 91 & 98 & 84 & \textbf{100} & 97 & 98 & \underline{99}
        \\

        & & Correct (\%)
        & \underline{74} & 28 & 65 & 63 & 40 & 40 & 17 & 41 & 70 & 56 & \textbf{78}
        \\

        & & Fast (\%)
        & \underline{40} & 17 & 14 & 15 & 25 & 30 & 7 & 24 & 37 & 29 & \textbf{47}
        \\

        \cmidrule(lr){2-14}

        & \multirow{3}{*}{L3}
        & Compiled (\%)
        & 90 & 86 & 60 & 60 & 86 & 90 & 82 & \textbf{100} & 84 & \underline{98} & \textbf{100}
        \\

        & & Correct (\%)
        & 34 & 12 & 30 & 20 & 4 & 30 & 18 & 38 & \underline{40} & 18 & \textbf{58}
        \\

        & & Fast (\%)
        & 10 & 4 & \textbf{14} & \underline{12} & 2 & 0 & 0 & 6 & 8 & 2 & \textbf{14}
        \\

        \midrule

        TritonBench-G
        & \multicolumn{2}{l}{Correctness pass@1 (\%)}
        & \textbf{20.1} & 11.4 & 17.9 & 18.5 & 19.0 & 19.0 & 19.0 & \underline{19.5} & \textbf{20.1} & \underline{19.5} & 14.9
        \\

        \bottomrule
    \end{tabular}
    }

    \begin{minipage}{\linewidth}
        \vspace{0.4em}
        \footnotesize
        KernelBench reports compiled, correct, and fast results at pass@1.
        L1 and L2 each contain 100 tasks; L3 contains 50 tasks, so its counts
        are converted to percentages (e.g., 45/17/5 becomes 90/34/10).
    \end{minipage}
\end{table}

\paratitle{RTL Performance Comparison}
Table~\ref{tab:industrial-coding-results} shows that \ourmethod is consistently
competitive across the RTL suite, placing among the top three systems on four
of the seven reported metrics. It achieves the best RTLLM functional avg@4 at
68.0, ranks second on VerilogEval code-completion avg@4 at 86.0 and CVDP
functional avg@5 at 44.1, and ranks third on VerilogEval spec-to-RTL avg@4 at
86.3. On the more complex benchmarks where it does not lead, its 61.7 RealBench
syntax pass@5 exceeds both InCoder variants, DeepSeek V4-Pro, GLM 5.2, Kimi
K2.6, and Hy3, while its 49.3 ArchXBench pass@1 remains above both InCoder
variants, Kimi K2.6, and Hy3. The clearest gain over the Qwen3.6-27B base model
appears on RTLLM, where post-training raises performance from 49.6 to 68.0, an
absolute improvement of 18.4 points.

\paratitle{Kernel Performance Comparison}
\ourmethod likewise ranks among the top three systems on seven of the nine
KernelBench metrics and shares the best TritonBench-G correctness pass@1 of
20.1. On KernelBench, it attains the highest L1 correctness rate at 61\%, then
ranks second on both L2 correctness and fast rates at 74\% and 40\%,
respectively. Its advantages remain visible on L3 despite the harder setting:
34\% correctness exceeds InCoder-32B, DeepSeek V4-Pro, GLM 5.2, Kimi K2.6, and
Hy3, while the 10\% fast rate exceeds GPT 5.5 and Claude Opus-4.8. Together,
these results place \ourmethod at or near the front of both operator-level and
repository-derived kernel generation while retaining competitive performance
on full-network workloads.

\subsection{RLVR Training Dynamics}

\begin{figure}[!t]
    \centering
    \includegraphics[width=\linewidth]{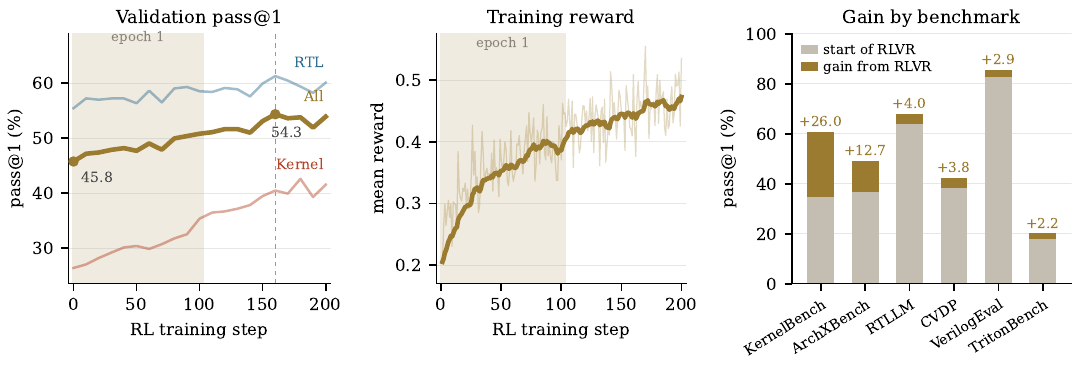}
    \caption{RLVR training dynamics and outcome. \emph{All} is the unweighted
    mean of the six verification backends, and \emph{Kernel} and \emph{RTL} the
    means within each domain; the bold reward curve is an exponential moving
    average of the per-step values. Shaded bands mark the first epoch and the
    dashed line the reported checkpoint.}
    \label{fig:rlvr-analysis}
\end{figure}

\paratitle{From reward to accuracy.}
The left and centre panels of Figure~\ref{fig:rlvr-analysis} report validation
pass@1 and mean reward per rollout over the run. Both rise together throughout:
reward from 0.295 at
step 20 to 0.446 at step 160, and the unweighted mean of the six rates from
45.8\% to 54.3\%.
The movement is concentrated early, with roughly the first hundred steps
accounting for 59\% of the aggregate gain, three quarters of the gain on
KernelBench, and all of the gain on ArchXBench. Mean reward continues to 0.473
at step 200 with no inflection toward saturation. Execution reward is therefore
converted into pass@1 within roughly one epoch.

\paratitle{Gains on complex designs.}
The right panel shows that gain varies across benchmarks, and it is
largest on those with the most complex designs. KernelBench, which asks for a
complete optimised implementation and spans single operators, operator fusion,
and whole network architectures, rises 26.0 points, and ArchXBench, which covers
architectural hardware designs, rises 12.7. VerilogEval and RTLLM, whose
problems are shorter and simpler and on which the model already scored 82.7\%
and 64.0\%, gain 2.9 and 4.0. The largest RLVR gains therefore occur on
demanding designs with greater improvement headroom.

\paratitle{Cross-domain compatibility.}
All six backends improve under a single reward interface and a kernel-to-RTL
ratio near one. Gains are largest on kernels, and every RTL backend also
improves, indicating cross-domain compatibility under shared execution reward.

\subsection{Case study.}

\paratitle{Before v.s. After RLVR} Figure~\ref{fig:rlvr-case-study} pairs two validation problems with the model's
own reasoning before and after RLVR: a Frobenius-norm normalisation from
KernelBench, and a 32-bit pipelined ripple-carry adder from level 2 of
ArchXBench. On the kernel task both versions reach the same dead end, and the
difference is what follows it. Before RLVR the model restates that dead end
until its budget is gone; after RLVR it accepts the conclusion and routes around
it with a second kernel. On the RTL task the version before RLVR gets every stage of
the arithmetic right and the pipeline skew wrong, while the version after RLVR
states the alignment rule and builds the delay chain it implies. These failures
concern recovery from a dead end and alignment across pipeline stages. Both
require coordination across operations, consistent with the larger RLVR gains
on complex designs.

\newtcolorbox{rlvrbefore}[1][]{
    enhanced, sharp corners, boxrule=0.4pt, left=4pt, right=4pt, top=4pt,
    bottom=4pt, colback=black!3, colframe=black!20,
    fonttitle=\footnotesize\bfseries, coltitle=black!58,
    attach boxed title to top left={xshift=7pt, yshift*=-\tcboxedtitleheight/2},
    boxed title style={sharp corners, boxrule=0pt, colback=white, left=3pt,
                       right=3pt, top=0pt, bottom=0pt},
    title={Before RLVR}, #1}

\newtcolorbox{rlvrafter}[1][]{
    enhanced, sharp corners, boxrule=0.4pt, left=4pt, right=4pt, top=4pt,
    bottom=4pt, colback=goldbg, colframe=goldaccent!50,
    fonttitle=\footnotesize\bfseries, coltitle=golddark,
    attach boxed title to top left={xshift=7pt, yshift*=-\tcboxedtitleheight/2},
    boxed title style={sharp corners, boxrule=0pt, colback=white, left=3pt,
                       right=3pt, top=0pt, bottom=0pt},
    title={After RLVR}, #1}

\newcommand{\rlvrcase}[4]{
    \noindent{\small\bfseries #1\par}
    \vspace{2pt}
    \noindent
    \begin{minipage}[t]{0.487\linewidth}
        \begin{rlvrbefore}[equal height group=#1]\footnotesize\itshape\sloppy #2\end{rlvrbefore}
    \end{minipage}\hfill
    \begin{minipage}[t]{0.487\linewidth}
        \begin{rlvrafter}[equal height group=#1]\footnotesize\itshape\sloppy #3\end{rlvrafter}
    \end{minipage}\par
    \vspace{3pt}
    \noindent{\footnotesize #4\par}}

\begin{figure}[H]

    \rlvrcase{Kernel}
    {``\dots then the same kernel does the division after a global sync? Not
      possible. \dots then the same kernel does the division after a global
      sync? Not possible. \dots then the same kernel does the division after a
      global sync? Not possible. \dots''}
    {``\dots we need to wait for all blocks to finish before division. But
      Triton kernels don't have a global sync. So we can't do the division in
      the same kernel after the atomic adds \dots We could launch a second
      kernel.''}
    {The same dead end is reached in both; only after RLVR does the model leave
     it. \textcolor{golddark}{Restated 971 times with no kernel emitted
     $\rightarrow$ 63\% shorter, and correct.}}

    \vspace{5pt}

    \rlvrcase{RTL}
    {``\dots the partial sums \dots can be computed and then just passed
      through pipeline registers to the output \dots We can design a
      shift-register-like pipeline \dots reg [7:0] sum\_low2 \dots
      reg [15:0] sum\_low3 \dots reg [23:0] sum\_low4''}
    {``\dots the carry-in is delayed by one cycle per stage, so the A and B
      slices for that stage must also be delayed by the same number of cycles
      \dots we can just assign sum = \{sum4\_reg, sum3\_delay1, sum2\_delay2,
      sum1\_delay3\}''}
    {Both state the pass-through idea; only after RLVR is a cycle count
     attached to it. \textcolor{golddark}{Results one cycle late,
     2 of 50 tests $\rightarrow$ 50 of 50.}}

    \caption{Matched trajectories on two validation problems, sampled from the
    same checkpoints before and after RLVR. The excerpts reproduce model
    reasoning verbatim, and RTL verdicts come from the benchmark testbench.}
    \label{fig:rlvr-case-study}
\end{figure}

\paratitle{EDA as long-horizon RTL optimization.}
We next test whether the trained model can operate inside an iterative design
loop rather than emit a single answer. We adapt eight designs from RTLScout
\citep{arnold2026rtlscout} to the evaluation-driven propose--evaluate--refine
interface of SimpleTES \citep{ye2026structured}. The set comprises two
specification-driven synthesis tasks, five RTL-rewriting tasks, and one wide
finite-state machine. Starting from the run-local seed for each task, a model
proposes three revisions per iteration for twenty iterations. Each candidate
must pass the task's self-checking testbench and synthesize successfully before
it is ranked by technology-independent post-synthesis Yosys cell count. Lower
cell count is better; the search score is $1000/N_{\mathrm{cell}}$. Candidates
that evade accounting through black boxes, hierarchy preservation, or
multi-module decomposition are excluded. The supplied logs do not record
formal-equivalence checks, so we describe the retained candidates as
testbench-validated rather than formally equivalent.

\begin{figure}[!t]
    \centering
    \includegraphics[width=\linewidth]{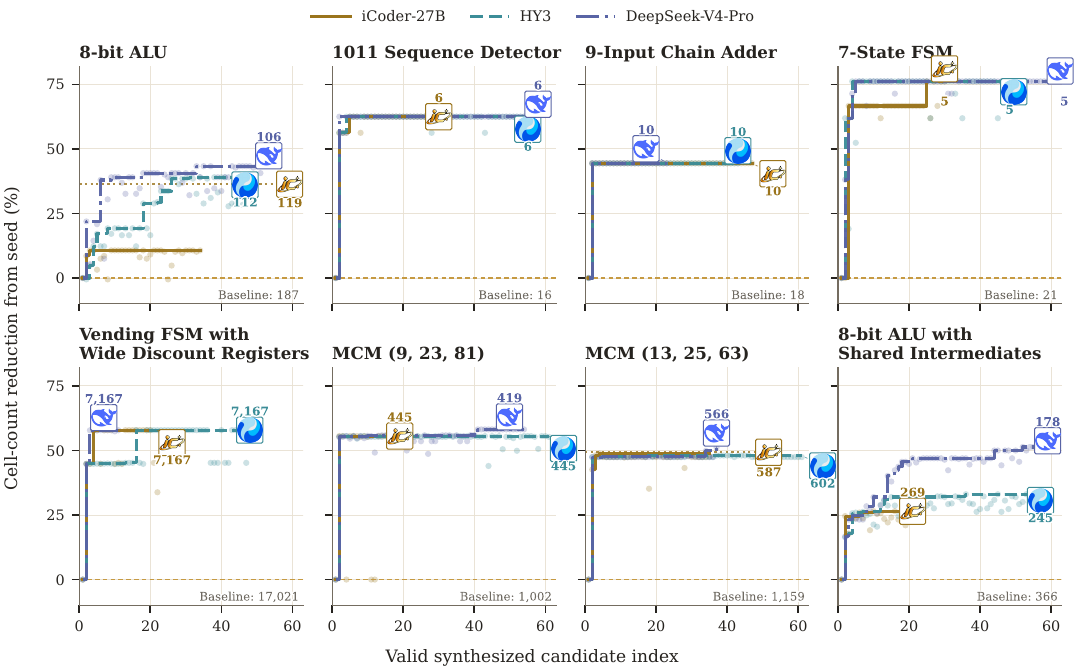}
    \caption{Evaluation-driven RTL search on eight designs adapted from
    RTLScout. Translucent points show every testbench-valid, successfully
    synthesized candidate retained by the evaluator; the opaque staircase is
    the cumulative best. Both are expressed as cell-count reduction from that
    task's recorded seed, and endpoint icons identify the model. Bold numbers
    beside the icons give the final synthesized cell counts; each panel reports
    its baseline cell count along the bottom. Dotted gold references on the
    8-bit ALU and MCM $(13,25,63)$ show refreshed iCoder final results whose
    candidate indices are unavailable. Their horizontal icon positions are
    visual anchors rather than candidate indices. Elsewhere, the horizontal
    axis counts valid synthesized candidates, not raw attempts or wall-clock
    rounds; failures and excluded candidates therefore do not advance it.}
    \label{fig:eda-search-trajectories}
\end{figure}

Figure~\ref{fig:eda-search-trajectories} shows every retained valid candidate
and the cumulative best, normalized to each task's seed. Its horizontal axis
counts valid synthesized candidates, because the records do not retain a
complete attempt-to-round mapping. The refreshed iCoder export reports final
solutions without the corresponding candidate indices; the two improved final
values are therefore shown as dotted gold references rather than appended to
the logged trajectory. With these updated results, iCoder reaches the common
best cell count on the 1011 sequence detector, 9-input registered chain adder,
7-state FSM, and vending FSM with wide discount registers. It ties HY3 on MCM
$(9,23,81)$ and improves on HY3 for MCM $(13,25,63)$.
Relative to HY3, this gives one win, five ties, and two losses. DeepSeek-V4-Pro
obtains a lower cell count on the remaining four tasks and ties on the other
four.

\begin{figure}[!t]
    \centering
    \includegraphics[width=\linewidth]{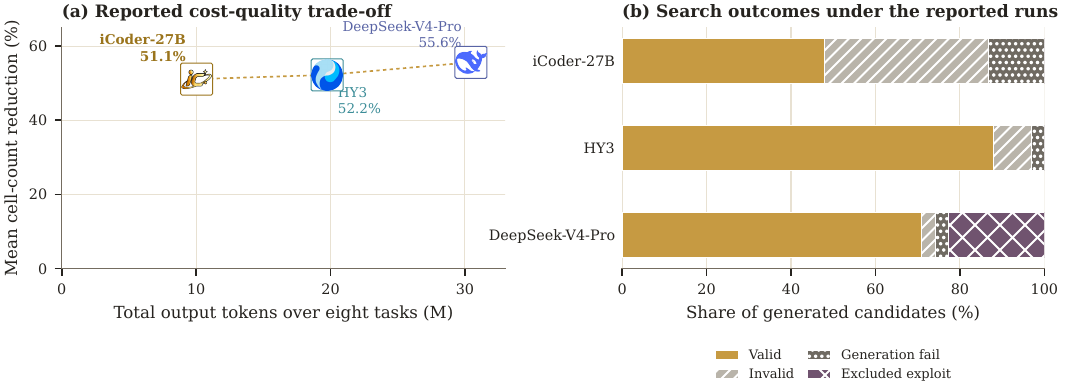}
    \caption{Reported cost--quality trade-off and search outcomes for the EDA
    case study. (a) Mean quality first computes cell-count reduction against
    the common recorded seed for each task and then averages the eight
    reductions with equal weight. Output tokens are summed across the recorded
    calls and serve as a cost proxy rather than monetary cost. (b) Outcome
    shares use the available candidate logs, retain all generated candidates in
    the denominator, and show reward-exploit candidates as a separate excluded
    category; the refreshed final-result metadata does not alter these counts.
    Runs nominally use 60 proposals per task; one DeepSeek-V4-Pro log contains
    an additional proposal. iCoder used a 32,768-token output cap, whereas HY3
    and DeepSeek-V4-Pro used 131,072.}
    \label{fig:eda-cost-quality-summary}
\end{figure}

Across tasks, we first compute the cell-count reduction relative to each
task's seed and then average the eight percentages with equal weight. This
avoids allowing small-cell designs to dominate the aggregate. As shown in
Figure~\ref{fig:eda-cost-quality-summary}, the mean reductions are 51.1\% for
iCoder, 52.2\% for HY3, and 55.6\% for DeepSeek-V4-Pro, leaving gaps of 1.1 and
4.5 percentage points. Their total output-token counts are 10.03M, 19.75M, and
30.45M, respectively. Thus, iCoder comes within 1.1 points of HY3 while using
51\% as many output tokens, and within 4.5 points of DeepSeek-V4-Pro while using
33\% as many.

The comparison is an exploratory case study rather than a controlled
token-matched benchmark. iCoder used a 32,768-token output cap, whereas HY3 and
DeepSeek-V4-Pro used 131,072; this difference may contribute to iCoder's larger
share of incomplete or invalid candidates. The metadata also omits token usage
for generation failures and does not retain complete candidate histories for
the refreshed iCoder final results, the Yosys version, or the complete synthesis
command. We therefore treat output tokens as a reported cost proxy, normalize
against the recorded seed within each run, and make no claim about dollar cost,
wall-clock time, or statistical significance.

\paratitle{GPU-kernel optimization under executable feedback.}
We next evaluate the three models in the SimpleTES propose--evaluate--refine loop
\citep{ye2026structured} for GPU-operator optimization. The archived suite
contains custom tasks with an RTLScout-inspired iterative-search scaffold
\citep{arnold2026rtlscout} and samples from the KernelFactory suite supplied
with the archive. Table~\ref{tab:kernel-case-study-operators} records every
operator, its computation, and the archive-designated reference implementation.
A candidate enters the search trace only when the archive records successful
compilation, correctness, and latency measurement.

\begin{figure}[!tbp]
    \centering
    \includegraphics[width=\linewidth]{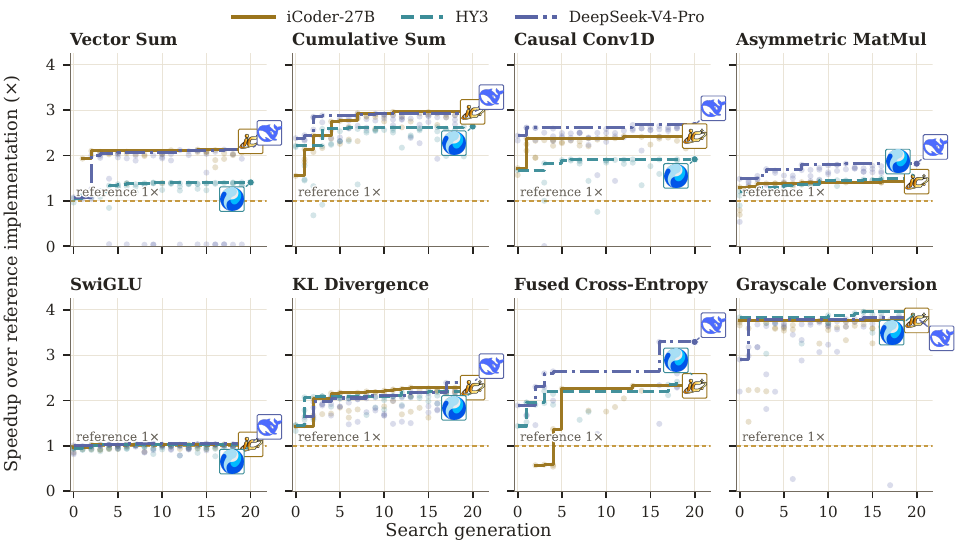}
    \caption{\textbf{Evaluation-driven search trajectories for eight GPU
    operators.} Translucent points are distinct generation-indexed candidates
    that the archive marks as compiled and correctness-valid; exact duplicate
    records are removed. The opaque staircase is the cumulative best. Every
    latency is divided into the same task's archive-designated reference
    latency, and the dashed line marks the resulting $1\times$ anchor. Endpoint icons identify
    the model. When endpoints overlap, a short leader connects an offset icon
    card to its exact point. Candidates without a recorded search-generation
    identifier do not appear in this plot.}
    \label{fig:kernel-search-trajectories}
\end{figure}

\begin{figure}[!tbp]
    \centering
    \includegraphics[width=\linewidth]{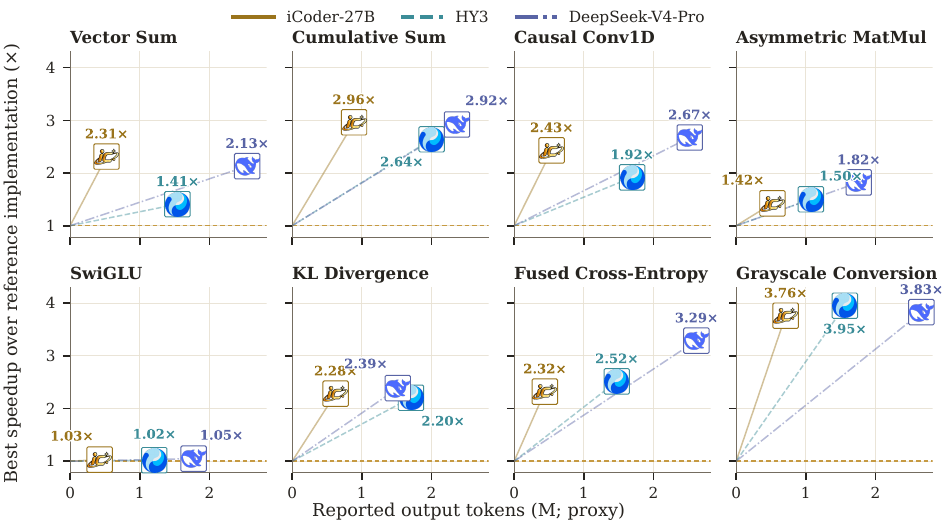}
    \caption{\textbf{Reported cost--performance trade-off by operator.} Each
    endpoint pairs the run's total reported output tokens with its best
    reference-relative speedup. A ray begins at the archive-designated reference
    implementation $(0,1)$, so its slope visualizes the descriptive gain per
    million reported output tokens used for case selection. Tokenizers, output limits, and proposal
    counts are not matched across models; the horizontal axis is a search-cost
    proxy, not dollar cost, GPU time, or controlled compute.}
    \label{fig:kernel-cost-performance}
\end{figure}

The source records mix naive-relative and reference-relative speedups. We
recompute every endpoint as
$s^{\mathrm{ref}}_{m,t}=\tau^{\mathrm{ref}}_t/\tau_{m,t}$, where
$\tau^{\mathrm{ref}}_t$ is the archived reference latency for task $t$. For a
compact $2\times4$ display, we use the post-hoc statistic
$g_{m,t}=(s^{\mathrm{ref}}_{m,t}-1)/\widetilde T_{m,t}$, with
$\widetilde T_{m,t}=T_{m,t}/10^6$ denoting millions of reported output tokens,
and rank complete three-model tasks by
$r_t=g_{\mathrm{iCoder},t}/\max\{g_{\mathrm{HY3},t},
g_{\mathrm{DeepSeek},t}\}$. We retain the eight largest $r_t$ values. GDNO is
omitted because HY3 has no valid endpoint; Histogram and Softmax receive the
two lowest $r_t$ values among complete tasks. This archive-token proxy is used
only to select and order illustrative panels. Figures~\ref{fig:kernel-search-trajectories} and
\ref{fig:kernel-cost-performance} show the retained search records and their
cost--performance endpoints.

iCoder obtains the highest reported speedup on Vector Sum ($2.31\times$) and
Cumulative Sum ($2.96\times$). The best archived iCoder endpoint is within
1.7\% of the fastest SwiGLU result, 4.4\% on KL Divergence, 4.8\% on Grayscale
Conversion, and 9.1\% on Causal Conv1D. The gaps reach 21.8\% on Asymmetric
MatMul and 29.4\% on Fused Cross-Entropy. Within this selected subset, the
descriptive geometric-mean reference-relative speedups are $2.17\times$,
$1.98\times$, and $2.36\times$ for iCoder, HY3, and DeepSeek-V4-Pro. Their
recorded output-token totals are 4.67M, 12.26M, and 17.75M. By construction,
$r_t$ ranges from $1.74$ to $5.65$ across the displayed subset; we do not
interpret this range as model-level efficiency.

These results do not establish budget-matched or hardware-controlled
efficiency. Proposal counts differ, and tokenizer identities, output limits,
the GPU model, and software versions are not recorded in the supplied kernel
archive. The package also omits raw verifier code, timing repetitions, and
uncertainty estimates. The figures report archived endpoints and a token proxy
without claims about monetary cost, GPU time, or statistical significance, and
the selected-subset aggregate does not characterize the full eleven-task
archive.

\FloatBarrier

\section{Conclusion}
\label{sec:conclusion}

Can recursive AI truly close the loop in which AI trains and improves AI? We
believe that this crown jewel remains beyond immediate reach. Rather than
claiming fully autonomous recursive self-improvement, this work presents a
practical path in which human experts and an AI agent jointly develop a frontier-competitive industrial-coding model. Human experts provide
executable research skills that define the objective, stage scaffold,
permission boundaries, verifier integrity, and operating procedures, while
leaving empirical choices to the agent. Within this trusted space, the agent
evolved executable data, selected capability gaps for SFT, explored and
revised OPSD designs, and iterated RLVR rewards and training constraints using
executable feedback, ultimately producing the 27B \ourmethod model. \ourmethod
achieves 68.0 on RTLLM and 61\% KernelBench L1 correctness, surpassing much
larger models including DeepSeek-V4-Pro, GLM-5.2, and Kimi-K2.6, as well as
GPT-5.5, Claude Opus 4.8, and Gemini 3.5 Flash. It also ties the best result on
TritonBench-G at 20.1 and ranks second on CVDP at 44.1. More broadly, this
division of labor offers an auditable path for AI for AI, where human expertise
defines a safe research space and agents accumulate evidence, revise
hypotheses, and improve models. We hope it helps move recursive AI toward
systems in which AI can assume an increasingly complete role in training the
next generation of AI.
\section*{Acknowledgments}

We sincerely thank Jiayi Zhang for contributions to the theoretical development
of our OPSD prior and for guidance on the autoresearch framework.

\begingroup
\renewcommand{\bibfont}{\footnotesize}
\bibliographystyle{assets/plainnat}
\bibliography{references}

\begin{thebibliography}{49}
\providecommand{\natexlab}[1]{#1}
\providecommand{\url}[1]{\texttt{#1}}
\expandafter\ifx\csname urlstyle\endcsname\relax
  \providecommand{\doi}[1]{doi: #1}\else
  \providecommand{\doi}{doi: \begingroup \urlstyle{rm}\Url}\fi

\bibitem[Acikgoz et~al.(2025)Acikgoz, Qian, Ji, Hakkani-T{\"u}r, and Tur]{acikgoz2025selfimproving}
Emre~Can Acikgoz, Cheng Qian, Heng Ji, Dilek Hakkani-T{\"u}r, and Gokhan Tur.
\newblock Self-improving {LLM} agents at test-time.
\newblock \emph{arXiv preprint arXiv:2510.07841}, 2025.
\newblock \doi{10.48550/arXiv.2510.07841}.
\newblock \url{https://arxiv.org/abs/2510.07841}.

\bibitem[Agarwal et~al.(2024)Agarwal, Vieillard, Zhou, Stanczyk, Ramos, Geist, and Bachem]{agarwal2024gkd}
Rishabh Agarwal, Nino Vieillard, Yongchao Zhou, Piotr Stanczyk, Sabela Ramos, Matthieu Geist, and Olivier Bachem.
\newblock On-policy distillation of language models: Learning from self-generated mistakes.
\newblock In \emph{International Conference on Learning Representations}, 2024.
\newblock \doi{10.48550/arXiv.2306.13649}.

\bibitem[Alzubi et~al.(2026)Alzubi, Provenzano, Bingham, Chen, and Vu]{alzubi2026evoskill}
Salaheddin Alzubi, Noah Provenzano, Jaydon Bingham, Weiyuan Chen, and Tu~Vu.
\newblock {EvoSkill}: Automated skill discovery for multi-agent systems.
\newblock \emph{arXiv preprint arXiv:2603.02766}, 2026.
\newblock \doi{10.48550/arXiv.2603.02766}.
\newblock \url{https://arxiv.org/abs/2603.02766}.

\bibitem[{Anthropic}(2026)]{anthropic2026opus48}
{Anthropic}.
\newblock Introducing {Claude Opus 4.8}.
\newblock Anthropic Product Release, May 2026.
\newblock \url{https://www.anthropic.com/news/claude-opus-4-8}.

\bibitem[Arnold et~al.(2026)Arnold, Amaudruz, Tsaras, Andri, and Cavigelli]{arnold2026rtlscout}
Felix Arnold, Ryan Amaudruz, Dimitrios Tsaras, Renzo Andri, and Lukas Cavigelli.
\newblock {RTLScout}: Joint agentic code and synthesis optimization for efficient digital circuits.
\newblock \emph{arXiv preprint arXiv:2606.06530}, 2026.
\newblock \doi{10.48550/arXiv.2606.06530}.
\newblock \url{https://arxiv.org/abs/2606.06530}.

\bibitem[Assump{\c{c}}{\~a}o et~al.(2025)Assump{\c{c}}{\~a}o, Ferreira, Campos, and Murai]{assumpcao2025codeevolve}
Henrique Assump{\c{c}}{\~a}o, Diego Ferreira, Leandro Campos, and Fabricio Murai.
\newblock Codeevolve: an open source evolutionary coding agent for algorithmic discovery and optimization.
\newblock \emph{arXiv preprint arXiv:2510.14150}, 2025.

\bibitem[Brodsky et~al.(2026)Brodsky, Kumar, Kashmira, Danatanarayana, Mars, Flautner, and Tang]{brodsky2026kernelforge}
Joshua Brodsky, Dhravid Kumar, Savini Kashmira, Jayanaka Danatanarayana, Jason Mars, Krisztian Flautner, and Lingjia Tang.
\newblock Kernel forge: An agent harness for {LLM}-based generation and optimization of {CUDA} kernels.
\newblock \emph{arXiv preprint arXiv:2607.24762}, 2026.
\newblock \doi{10.48550/arXiv.2607.24762}.
\newblock \url{https://arxiv.org/abs/2607.24762}.

\bibitem[Chen et~al.(2026)Chen, Wang, and Qu]{chen2026rsi}
Mingguang Chen, Licheng Wang, and Bo~Qu.
\newblock Recursive self-improvement in {AI}: From bounded self-refinement to autonomous research loops.
\newblock \emph{arXiv preprint arXiv:2607.07663}, 2026.
\newblock \doi{10.48550/arXiv.2607.07663}.
\newblock \url{https://arxiv.org/abs/2607.07663}.

\bibitem[{DeepSeek-AI}(2026)]{deepseek2026v4}
{DeepSeek-AI}.
\newblock {DeepSeek-V4} preview release.
\newblock DeepSeek API Documentation, April 2026.
\newblock \url{https://api-docs.deepseek.com/news/news260424/}.

\bibitem[Deng et~al.(2026)Deng, Yu, Liu, Pinckney, Khailany, and Ren]{deng2026acertl}
Chenhui Deng, Zhongzhi Yu, Guan-Ting Liu, Nathaniel Pinckney, Brucek Khailany, and Haoxing Ren.
\newblock {ACE-RTL}: When agentic context evolution meets {RTL}-specialized {LLMs}.
\newblock \emph{arXiv preprint arXiv:2602.10218}, 2026.
\newblock \doi{10.48550/arXiv.2602.10218}.
\newblock \url{https://arxiv.org/abs/2602.10218}.

\bibitem[Ding et~al.(2026)Ding, Nannapaneni, Liu, and Zhang]{ding2026verificationgap}
Tianyu Ding, Aditya Nannapaneni, Bingfan Liu, and Ling Zhang.
\newblock Autonomous research agents: A survey of {AI} scientists and the verification gap.
\newblock \emph{arXiv preprint arXiv:2608.05179}, 2026.
\newblock \doi{10.48550/arXiv.2608.05179}.
\newblock \url{https://arxiv.org/abs/2608.05179}.

\bibitem[Du et~al.(2026)Du, Yan, Shi, Cao, Feng, Liang, Sun, Peng, Zhou, Li, Zhou, He, Zhang, and Bai]{du2026mlevolve}
Shangheng Du, Xiangchao Yan, Jinxin Shi, Zongsheng Cao, Shiyang Feng, Zichen Liang, Boyuan Sun, Tianshuo Peng, Yifan Zhou, Xin Li, Jie Zhou, Liang He, Bo~Zhang, and Lei Bai.
\newblock {MLEvolve}: A self-evolving framework for automated machine learning algorithm discovery.
\newblock \emph{arXiv preprint arXiv:2606.06473}, 2026.
\newblock \doi{10.48550/arXiv.2606.06473}.
\newblock \url{https://arxiv.org/abs/2606.06473}.

\bibitem[Fu et~al.(2026)Fu, Wang, Shao, Karri, Shafique, Knechtel, Sinanoglu, and Guo]{fu2026synthesis}
Weimin Fu, Zeng Wang, Minghao Shao, Ramesh Karri, Muhammad Shafique, Johann Knechtel, Ozgur Sinanoglu, and Xiaolong Guo.
\newblock Synthesis-in-the-loop evaluation of {LLMs} for {RTL} generation: Quality, reliability, and failure modes.
\newblock \emph{arXiv preprint arXiv:2603.11287}, 2026.
\newblock \doi{10.48550/arXiv.2603.11287}.
\newblock \url{https://arxiv.org/abs/2603.11287}.

\bibitem[{Google DeepMind}(2026)]{google2026gemini35}
{Google DeepMind}.
\newblock {Gemini 3.5 Flash} model card.
\newblock Google DeepMind Model Card, May 2026.
\newblock \url{https://deepmind.google/models/model-cards/gemini-3-5-flash/}.

\bibitem[Guo et~al.(2026)Guo, Huang, Gao, Li, Ge, Kuang, and Wang]{guo2026aqua}
Jiacheng Guo, Suozhi Huang, Yunlong Gao, Zihao Li, Jason Ge, Xu~Kuang, and Mengdi Wang.
\newblock {AQuA}: Recursively self-improving quantitative trading research agents.
\newblock \emph{arXiv preprint arXiv:2608.12841}, 2026.
\newblock \doi{10.48550/arXiv.2608.12841}.
\newblock \url{https://arxiv.org/abs/2608.12841}.

\bibitem[Huang et~al.(2026)Huang, Wen, Xu, Yan, Xia, and Liu]{huang2026realistictriton}
Jinjun Huang, Zhongzhen Wen, Tongtong Xu, Meng Yan, Xin Xia, and Zhongxin Liu.
\newblock {RealisticTritonBench}: A benchmark for {Triton}-kernel generation in real-world {AI} frameworks.
\newblock \emph{arXiv preprint arXiv:2608.12004}, 2026.
\newblock \doi{10.48550/arXiv.2608.12004}.
\newblock \url{https://arxiv.org/abs/2608.12004}.

\bibitem[H{\"u}botter et~al.(2026)H{\"u}botter, L{\"u}beck, Behric, Baumann, Bagatella, Marta, Hakimi, Shenfeld, Kleine~Buening, Guestrin, and Krause]{hubotter2026sdpo}
Jonas H{\"u}botter, Frederike L{\"u}beck, Lejs Behric, Anton Baumann, Marco Bagatella, Daniel Marta, Ido Hakimi, Idan Shenfeld, Thomas Kleine~Buening, Carlos Guestrin, and Andreas Krause.
\newblock Reinforcement learning via self-distillation.
\newblock \emph{arXiv preprint arXiv:2601.20802}, 2026.
\newblock \doi{10.48550/arXiv.2601.20802}.

\bibitem[Jaber and Jaber(2026)]{jaber2026autokernel}
Jaber Jaber and Osama Jaber.
\newblock {AutoKernel}: Autonomous {GPU} kernel optimization via iterative agent-driven search.
\newblock \emph{arXiv preprint arXiv:2603.21331}, 2026.
\newblock \doi{10.48550/arXiv.2603.21331}.
\newblock \url{https://arxiv.org/abs/2603.21331}.

\bibitem[Jin et~al.(2025)Jin, Huang, Li, Cheng, Zhao, Zheng, Zhu, Xing, Dou, Zhang, Du, Guo, and Hu]{jin2025realbench}
Pengwei Jin, Di~Huang, Chongxiao Li, Shuyao Cheng, Yang Zhao, Xinyao Zheng, Jiaguo Zhu, Shuyi Xing, Bohan Dou, Rui Zhang, Zidong Du, Qi~Guo, and Xing Hu.
\newblock {RealBench}: Benchmarking {Verilog} generation models with real-world {IP} designs.
\newblock \emph{arXiv preprint arXiv:2507.16200}, 2025.
\newblock \doi{10.48550/arXiv.2507.16200}.
\newblock \url{https://arxiv.org/abs/2507.16200}.

\bibitem[Lee et~al.(2026)Lee, Xu, Seely, Lee, Zaharia, and Tang]{lee2026rhi}
Hyunin Lee, Jinglue Xu, Jeffrey Seely, Donghyun Lee, Matei Zaharia, and Yujin Tang.
\newblock Recursive harness self-improvement.
\newblock \emph{arXiv preprint arXiv:2607.15524}, 2026.
\newblock \doi{10.48550/arXiv.2607.15524}.
\newblock \url{https://arxiv.org/abs/2607.15524}.

\bibitem[Li et~al.(2026)]{li2026trex}
Dongfang Li et~al.
\newblock {SLAI T-Rex}: Full-parameter post-training of the {DeepSeek-V4} family on {Ascend SuperPOD}.
\newblock \emph{arXiv preprint arXiv:2607.20145}, 2026.
\newblock \doi{10.48550/arXiv.2607.20145}.
\newblock \url{https://arxiv.org/abs/2607.20145}.

\bibitem[Li et~al.(2025)Li, Li, Gao, Shi, Li, Wang, Huang, Wang, Wang, Han, Liu, and Sun]{li2025tritonbench}
Jianling Li, Shangzhan Li, Zhenye Gao, Qi~Shi, Yuxuan Li, Zefan Wang, Jiacheng Huang, Haojie Wang, Jianrong Wang, Xu~Han, Zhiyuan Liu, and Maosong Sun.
\newblock {TritonBench}: Benchmarking large language model capabilities for generating {Triton} operators.
\newblock \emph{arXiv preprint arXiv:2502.14752}, 2025.
\newblock \doi{10.48550/arXiv.2502.14752}.
\newblock \url{https://arxiv.org/abs/2502.14752}.

\bibitem[Lin et~al.(2026)Lin, Liu, and Yan]{lin2026phynex}
Hang Lin, Chongwen Liu, and Gang Yan.
\newblock Large language model based agent for automated discovery in computational physics.
\newblock \emph{arXiv preprint arXiv:2606.14266}, 2026.
\newblock \doi{10.48550/arXiv.2606.14266}.
\newblock \url{https://arxiv.org/abs/2606.14266}.

\bibitem[Liu et~al.(2023)Liu, Pinckney, Khailany, and Ren]{liu2023verilogeval}
Mingjie Liu, Nathaniel Pinckney, Brucek Khailany, and Haoxing Ren.
\newblock {VerilogEval}: Evaluating large language models for {Verilog} code generation.
\newblock \emph{arXiv preprint arXiv:2309.07544}, 2023.
\newblock \doi{10.48550/arXiv.2309.07544}.
\newblock \url{https://arxiv.org/abs/2309.07544}.

\bibitem[Lu et~al.(2023)Lu, Liu, Zhang, and Xie]{lu2023rtllm}
Yao Lu, Shang Liu, Qijun Zhang, and Zhiyao Xie.
\newblock {RTLLM}: An open-source benchmark for design {RTL} generation with large language model.
\newblock \emph{arXiv preprint arXiv:2308.05345}, 2023.
\newblock \doi{10.48550/arXiv.2308.05345}.
\newblock \url{https://arxiv.org/abs/2308.05345}.

\bibitem[Luo(2026)]{luo2026xscientist}
Jixiang Luo.
\newblock {XScientist}: A git-like research protocol for long-running autonomous scientific discovery.
\newblock \emph{arXiv preprint arXiv:2607.12301}, 2026.
\newblock \doi{10.48550/arXiv.2607.12301}.
\newblock \url{https://arxiv.org/abs/2607.12301}.

\bibitem[Novikov et~al.(2025)Novikov, V{\~u}, Eisenberger, Dupont, Huang, Wagner, Shirobokov, Kozlovskii, Ruiz, Mehrabian, Kumar, See, Chaudhuri, Holland, Davies, Nowozin, Kohli, and Balog]{novikov2025alphaevolve}
Alexander Novikov, Ng\^an V{\~u}, Marvin Eisenberger, Emilien Dupont, Po-Sen Huang, Adam~Zsolt Wagner, Sergey Shirobokov, Borislav Kozlovskii, Francisco J.~R. Ruiz, Abbas Mehrabian, M.~Pawan Kumar, Abigail See, Swarat Chaudhuri, George Holland, Alex Davies, Sebastian Nowozin, Pushmeet Kohli, and Matej Balog.
\newblock {AlphaEvolve}: A coding agent for scientific and algorithmic discovery.
\newblock \emph{arXiv preprint arXiv:2506.13131}, 2025.
\newblock \doi{10.48550/arXiv.2506.13131}.
\newblock \url{https://arxiv.org/abs/2506.13131}.

\bibitem[{OpenAI}(2026)]{openai2026gpt55}
{OpenAI}.
\newblock Introducing {GPT-5.5}.
\newblock OpenAI Product Release, April 2026.
\newblock \url{https://openai.com/index/introducing-gpt-5-5/}.

\bibitem[Ouyang et~al.(2025)Ouyang, Guo, Arora, Zhang, Hu, R{\'e}, and Mirhoseini]{ouyang2025kernelbench}
Anne Ouyang, Simon Guo, Simran Arora, Alex~L. Zhang, William Hu, Christopher R{\'e}, and Azalia Mirhoseini.
\newblock {KernelBench}: Can {LLMs} write efficient {GPU} kernels?
\newblock \emph{arXiv preprint arXiv:2502.10517}, 2025.
\newblock \doi{10.48550/arXiv.2502.10517}.
\newblock \url{https://arxiv.org/abs/2502.10517}.

\bibitem[Pinckney et~al.(2025)Pinckney, Deng, Ho, Tsai, Liu, Zhou, Khailany, and Ren]{pinckney2025cvdp}
Nathaniel Pinckney, Chenhui Deng, Chia-Tung Ho, Yun-Da Tsai, Mingjie Liu, Wenfei Zhou, Brucek Khailany, and Haoxing Ren.
\newblock Comprehensive {Verilog} design problems: A next-generation benchmark dataset for evaluating large language models and agents on {RTL} design and verification.
\newblock \emph{arXiv preprint arXiv:2506.14074}, 2025.
\newblock \doi{10.48550/arXiv.2506.14074}.
\newblock \url{https://arxiv.org/abs/2506.14074}.

\bibitem[Purini et~al.(2025)Purini, Garg, Gaur, Bhat, Mupparapu, and Ravindran]{purini2025archxbench}
Suresh Purini, Siddhant Garg, Mudit Gaur, Sankalp Bhat, Sohan Mupparapu, and Arun Ravindran.
\newblock Archxbench: A complex digital systems benchmark suite for llm driven rtl synthesis.
\newblock In \emph{2025 ACM/IEEE 7th Symposium on Machine Learning for CAD (MLCAD)}, pages 1--10. IEEE, 2025.

\bibitem[{Qwen Team}(2026)]{qwen2026qwen36}
{Qwen Team}.
\newblock {Qwen3.6-27B}.
\newblock Qwen Blog, April 2026.
\newblock \url{https://qwen.ai/blog?id=qwen3.6-27b}.

\bibitem[Rank et~al.(2026)Rank, Bhatnagar, Prabhu, Eisenberg, Nguyen, Bethge, and Andriushchenko]{rank2026posttrainbench}
Ben Rank, Hardik Bhatnagar, Ameya Prabhu, Shira Eisenberg, Karina Nguyen, Matthias Bethge, and Maksym Andriushchenko.
\newblock {PostTrainBench}: Can {LLM} agents automate {LLM} post-training?
\newblock \emph{arXiv preprint arXiv:2603.08640}, 2026.
\newblock \doi{10.48550/arXiv.2603.08640}.
\newblock \url{https://arxiv.org/abs/2603.08640}.

\bibitem[Safdar and Saadeldin(2026)]{safdar2026longhorizon}
Aon Safdar and Mohamed Saadeldin.
\newblock Long-horizon autonomous architecture research with a language-model agent: A behavioural case study.
\newblock \emph{arXiv preprint arXiv:2608.01995}, 2026.
\newblock \doi{10.48550/arXiv.2608.01995}.
\newblock \url{https://arxiv.org/abs/2608.01995}.

\bibitem[Schulman et~al.(2017)Schulman, Wolski, Dhariwal, Radford, and Klimov]{schulman2017ppo}
John Schulman, Filip Wolski, Prafulla Dhariwal, Alec Radford, and Oleg Klimov.
\newblock Proximal policy optimization algorithms.
\newblock \emph{arXiv preprint arXiv:1707.06347}, 2017.
\newblock \doi{10.48550/arXiv.1707.06347}.

\bibitem[Shao et~al.(2024)Shao, Wang, Zhu, Xu, Song, Bi, Zhang, Zhang, Li, Wu, and Guo]{shao2024deepseekmath}
Zhihong Shao, Peiyi Wang, Qihao Zhu, Runxin Xu, Junxiao Song, Xiao Bi, Haowei Zhang, Mingchuan Zhang, Y.~K. Li, Y.~Wu, and Daya Guo.
\newblock {DeepSeekMath}: Pushing the limits of mathematical reasoning in open language models.
\newblock \emph{arXiv preprint arXiv:2402.03300}, 2024.
\newblock \doi{10.48550/arXiv.2402.03300}.

\bibitem[Sheng et~al.(2024)Sheng, Zhang, Ye, Wu, Zhang, Zhang, Peng, Lin, and Wu]{sheng2024hybridflow}
Guangming Sheng, Chi Zhang, Zilingfeng Ye, Xibin Wu, Wang Zhang, Ru~Zhang, Yanghua Peng, Haibin Lin, and Chuan Wu.
\newblock {HybridFlow}: A flexible and efficient {RLHF} framework.
\newblock \emph{arXiv preprint arXiv:2409.19256}, 2024.
\newblock \doi{10.48550/arXiv.2409.19256}.

\bibitem[Sutton(2019)]{sutton2019bitterlesson}
Rich Sutton.
\newblock The bitter lesson.
\newblock Incomplete Ideas, March 2019.
\newblock \url{http://www.incompleteideas.net/IncIdeas/BitterLesson.html}.
\newblock March 13, 2019.

\bibitem[Wang et~al.(2026{\natexlab{a}})Wang, Liu, Zhou, and Xie]{wang2026rtlbenchmt}
Jing Wang, Shang Liu, Hangan Zhou, and Zhiyao Xie.
\newblock {RTL-BenchMT}: Dynamic maintenance of {RTL} generation benchmark through agent-assisted analysis and revision.
\newblock \emph{arXiv preprint arXiv:2605.15537}, 2026{\natexlab{a}}.
\newblock \doi{10.48550/arXiv.2605.15537}.
\newblock \url{https://arxiv.org/abs/2605.15537}.

\bibitem[Wang et~al.(2026{\natexlab{b}})Wang, Yan, Bi, Yan, Tresp, and Ma]{wang2026metaskill}
Zefeng Wang, Minxi Yan, Jinhe Bi, Sikuan Yan, Volker Tresp, and Yunpu Ma.
\newblock {MetaSkill-Evolve}: Recursive self-improvement of {LLM} agents via two-timescale meta-skill evolution.
\newblock \emph{arXiv preprint arXiv:2607.05297}, 2026{\natexlab{b}}.
\newblock \doi{10.48550/arXiv.2607.05297}.
\newblock \url{https://arxiv.org/abs/2607.05297}.

\bibitem[Yamada et~al.(2026)Yamada, Lange, Lu, Lu, Hu, Foerster, Ha, and Clune]{yamada2026endtoend}
Yutaro Yamada, Robert~Tjarko Lange, Cong Lu, Chris Lu, Shengran Hu, Jakob Foerster, David Ha, and Jeff Clune.
\newblock Towards end-to-end automation of {AI} research.
\newblock \emph{arXiv preprint arXiv:2606.15497}, 2026.
\newblock \doi{10.48550/arXiv.2606.15497}.
\newblock \url{https://arxiv.org/abs/2606.15497}.

\bibitem[Yang et~al.(2026)Yang, Qin, Si, Chen, Gu, Yao, Lin, Wang, Wang, and Duan]{yang2026rlsd}
Chenxu Yang, Chuanyu Qin, Qingyi Si, Minghui Chen, Naibin Gu, Dingyu Yao, Zheng Lin, Weiping Wang, Jiaqi Wang, and Nan Duan.
\newblock Self-distilled {RLVR}.
\newblock \emph{arXiv preprint arXiv:2604.03128}, 2026.
\newblock \doi{10.48550/arXiv.2604.03128}.

\bibitem[Ye et~al.(2026)Ye, Lin, Tang, Luo, Thapa, Yang, Su, Yang, Liu, Li, Li, Sun, Gao, Ding, He, Zhang, Sun, Wang, Zhong, Shen, Li, Lu, Cui, He, Ma, Li, Baoyin, Choi, Ermon, Chu, Li, Xu, and Zou]{ye2026structured}
Haotian Ye, Haowei Lin, Jingyi Tang, Yizhen Luo, Rahul Thapa, Caiyin Yang, Chang Su, Rui Yang, Ruihua Liu, Rundao Li, Zeyu Li, Pengwei Sun, Chong Gao, Dachao Ding, Guangrong He, Miaolei Zhang, Lina Sun, Wenyang Wang, Yuchen Zhong, Zhuohao Shen, Puheng Li, Pan Lu, Bianxiao Cui, Di~He, Jianzhu Ma, Junfeng Li, Hexi Baoyin, Yejin Choi, Stefano Ermon, Xiaowen Chu, Tongyang Li, Yuzhi Xu, and James Zou.
\newblock Structured scaling of {AI} discovery across diverse scientific domains.
\newblock \emph{arXiv preprint arXiv:2604.19341}, 2026.
\newblock \doi{10.48550/arXiv.2604.19341}.
\newblock \url{https://arxiv.org/abs/2604.19341}.

\bibitem[Zang et~al.(2026)Zang, Tao, Zhang, Yuan, Zhang, Liu, and Lin]{zang2026kernelgenbench}
Peiyu Zang, Jian Tao, Jialing Zhang, Yichen Yuan, Wentao Zhang, Guang Liu, and Yonghua Lin.
\newblock Kernelgenbench: A multi-source and multi-chip benchmark for llm-based kernel generation.
\newblock \emph{arXiv preprint arXiv:2607.27231}, 2026.

\bibitem[Zhao et~al.(2026{\natexlab{a}})Zhao, Dong, Xu, Hasan, Fan, Jiang, Mao, {Ting Lingya}, Zou, Zhou, Chan, Zhang, Zhou, Huang, Li, Cun, Chen, Yuan, and Geng]{zhao2026scienceflow}
Mingming Zhao, Jiqian Dong, Kangping Xu, Zadid Hasan, Chengrui Fan, Shan Jiang, Shuai Mao, {Ting Lingya}, Linyi Zou, Tailin Zhou, Yun~Hin Chan, Wenkai Zhang, Zhanhong Zhou, Guowei Huang, Hongliang Li, Wenjing Cun, Zhitang Chen, Mingxuan Yuan, and Yanhui Geng.
\newblock {ScienceFlow}: A long-horizon agent for {ML} research, scientific discovery and beyond.
\newblock \emph{arXiv preprint arXiv:2608.14354}, 2026{\natexlab{a}}.
\newblock \doi{10.48550/arXiv.2608.14354}.
\newblock \url{https://arxiv.org/abs/2608.14354}.

\bibitem[Zhao et~al.(2026{\natexlab{b}})Zhao, Xie, Liu, Huang, Pang, Chen, and Grover]{zhao2026opsd}
Siyan Zhao, Zhihui Xie, Mengchen Liu, Jing Huang, Guan Pang, Feiyu Chen, and Aditya Grover.
\newblock Self-distilled reasoner: On-policy self-distillation for large language models.
\newblock \emph{arXiv preprint arXiv:2601.18734}, 2026{\natexlab{b}}.
\newblock \doi{10.48550/arXiv.2601.18734}.

\bibitem[Zheng et~al.(2025)Zheng, Liu, Li, Chen, Yu, Gao, Dang, Liu, Men, Yang, Zhou, and Lin]{zheng2025gspo}
Chujie Zheng, Shixuan Liu, Mingze Li, Xiong-Hui Chen, Bowen Yu, Chang Gao, Kai Dang, Yuqiong Liu, Rui Men, An~Yang, Jingren Zhou, and Junyang Lin.
\newblock Group sequence policy optimization.
\newblock \emph{arXiv preprint arXiv:2507.18071}, 2025.
\newblock \doi{10.48550/arXiv.2507.18071}.
\newblock \url{https://arxiv.org/abs/2507.18071}.

\bibitem[Zhou et~al.(2025)Zhou, Zhang, Tong, Zhang, Chen, Kong, Cai, Liu, Wang, Zhou, and Hoi]{zhou2025maiui}
Hanzhang Zhou, Xu~Zhang, Panrong Tong, Jianan Zhang, Liangyu Chen, Quyu Kong, Chenglin Cai, Chen Liu, Yue Wang, Jingren Zhou, and Steven Hoi.
\newblock {MAI-UI} technical report: Real-world centric foundation {GUI} agents.
\newblock \emph{arXiv preprint arXiv:2512.22047}, 2025.
\newblock \doi{10.48550/arXiv.2512.22047}.
\newblock \url{https://arxiv.org/abs/2512.22047}.

\bibitem[Zhou(2026)]{zhou2026hsi}
Tailin Zhou.
\newblock Hierarchical self-improvement: A framework for task-specific evolvable agent harnesses.
\newblock \emph{arXiv preprint arXiv:2608.08466}, 2026.
\newblock \doi{10.48550/arXiv.2608.08466}.
\newblock \url{https://arxiv.org/abs/2608.08466}.

\end{thebibliography}
\endgroup

\clearpage
\appendix
\section{Supplementary Material}
\label{app:supplementary-material}

\subsection{Task-Pool Distribution}
\label{app:data-analysis}

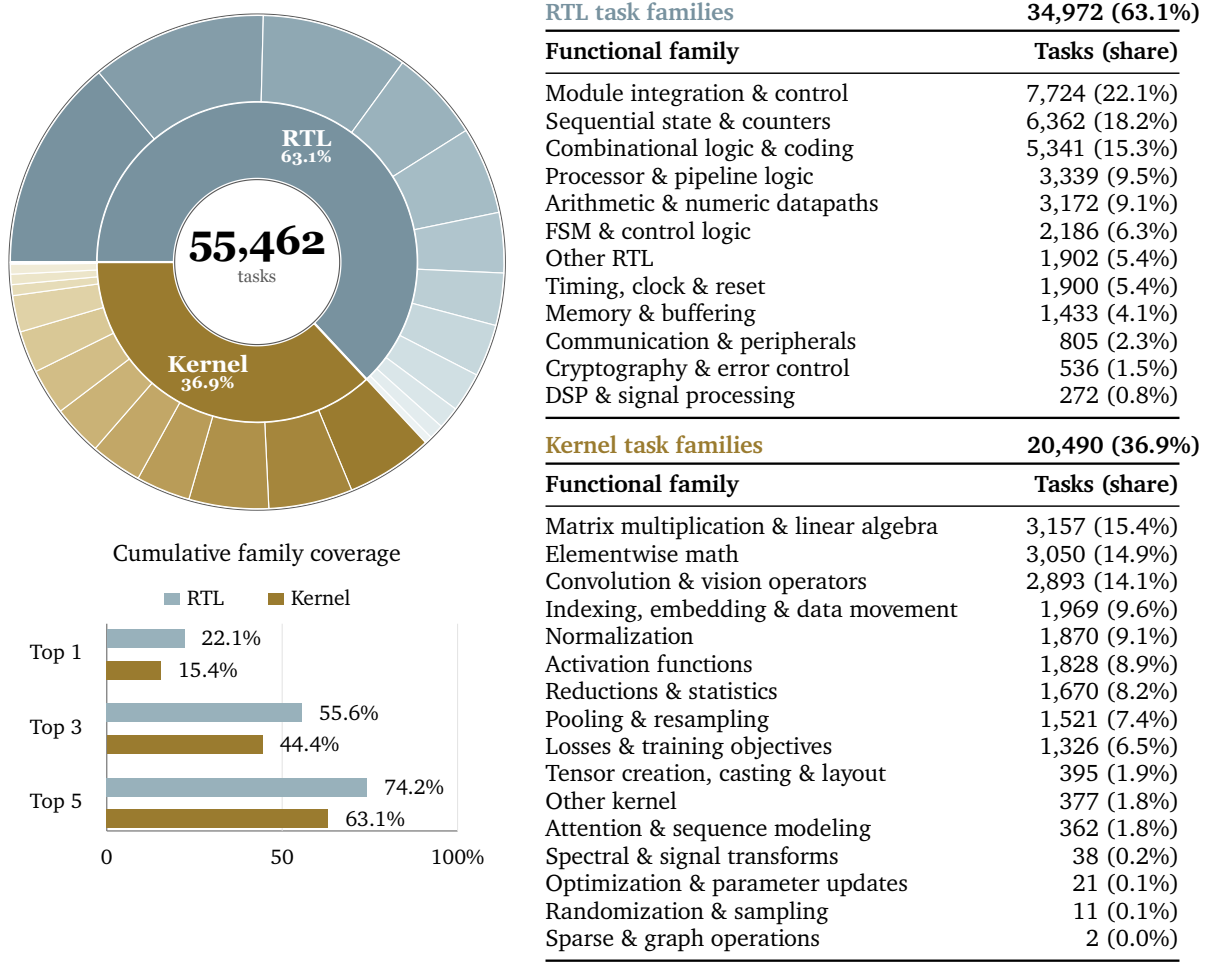
\begin{figure}[H]
  \centering
  \definecolor{taskpoolrtl}{HTML}{98B1BB}
\definecolor{taskpoolrtltext}{HTML}{78929F}
\definecolor{taskpoolkernel}{HTML}{9A7B2F}

\begin{minipage}[t]{0.43\linewidth}
  \vspace{0pt}
  \centering
  \includegraphics[width=\linewidth]{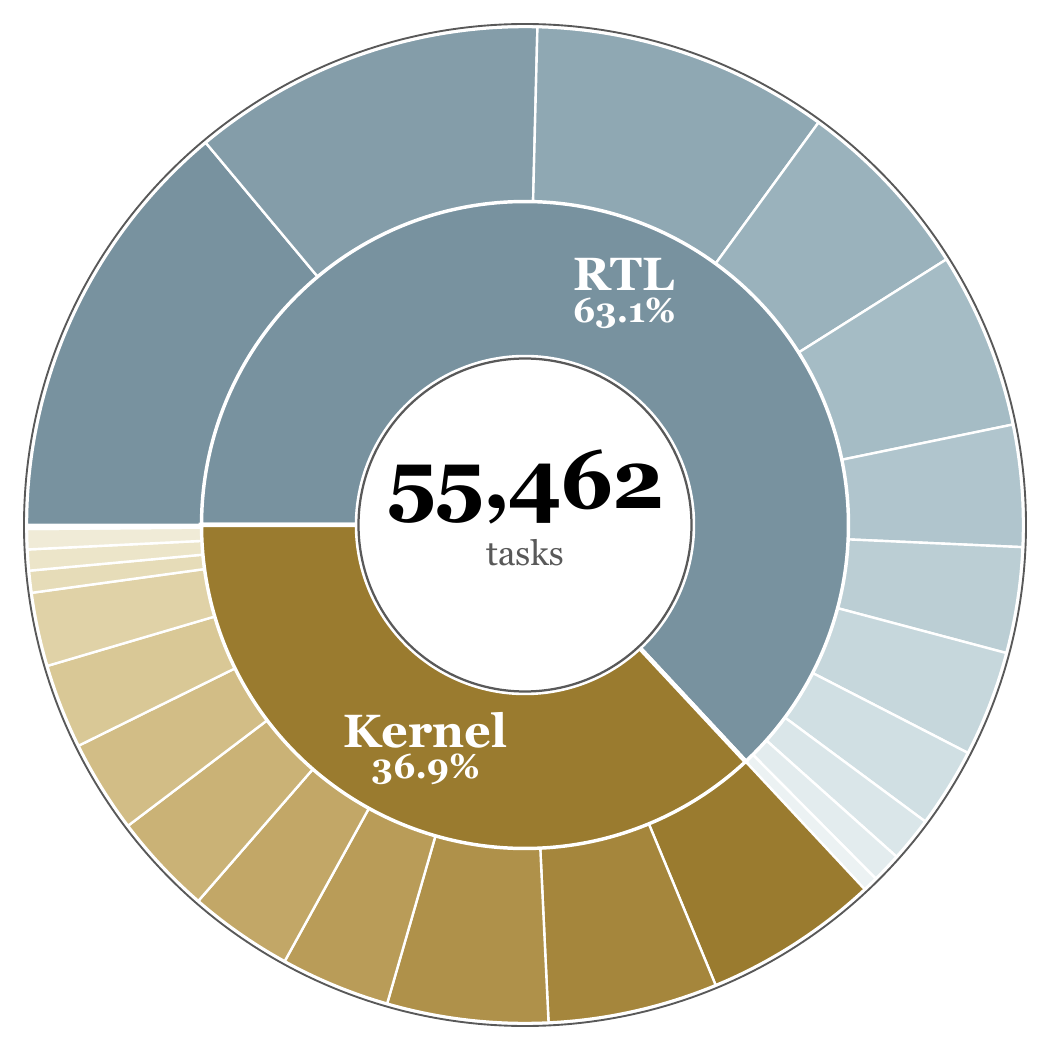}

  \vspace{7pt}
  \begin{minipage}{0.95\linewidth}
    \centering
    {\footnotesize Cumulative family coverage}\par
    \vspace{3pt}
    {\scriptsize
      \textcolor{taskpoolrtl}{\rule{0.75em}{0.55em}}\hspace{0.35em}RTL
      \hspace{1.5em}
      \textcolor{taskpoolkernel}{\rule{0.75em}{0.55em}}\hspace{0.35em}Kernel
    }\par
    \vspace{5pt}
    \begin{tikzpicture}[x=1.32pt,y=1pt]

      \draw[black!55,line width=0.45pt] (0,4) -- (0,82);
      \draw[black!55,line width=0.45pt] (0,4) -- (100,4);
      \draw[black!11,line width=0.35pt] (50,4) -- (50,82);
      \draw[black!11,line width=0.35pt] (100,4) -- (100,82);

      \node[anchor=east,font=\scriptsize] at (-4,70.5) {Top 1};
      \fill[taskpoolrtl] (0,73) rectangle (22.1,80);
      \fill[taskpoolkernel] (0,61) rectangle (15.4,68);
      \node[anchor=west,font=\scriptsize] at (24.1,76.5) {22.1\%};
      \node[anchor=west,font=\scriptsize] at (17.4,64.5) {15.4\%};

      \node[anchor=east,font=\scriptsize] at (-4,42.5) {Top 3};
      \fill[taskpoolrtl] (0,45) rectangle (55.6,52);
      \fill[taskpoolkernel] (0,33) rectangle (44.4,40);
      \node[anchor=west,font=\scriptsize] at (57.6,48.5) {55.6\%};
      \node[anchor=west,font=\scriptsize] at (46.4,36.5) {44.4\%};

      \node[anchor=east,font=\scriptsize] at (-4,14.5) {Top 5};
      \fill[taskpoolrtl] (0,17) rectangle (74.2,24);
      \fill[taskpoolkernel] (0,5) rectangle (63.1,12);
      \node[anchor=west,font=\scriptsize] at (76.2,20.5) {74.2\%};
      \node[anchor=west,font=\scriptsize] at (65.1,8.5) {63.1\%};

      \node[anchor=north,font=\scriptsize] at (0,1) {0};
      \node[anchor=north,font=\scriptsize] at (50,1) {50};
      \node[anchor=north,font=\scriptsize] at (100,1) {100\%};
    \end{tikzpicture}
  \end{minipage}
\end{minipage}\hfill
\begin{minipage}[t]{0.54\linewidth}
  \vspace{0pt}
  \footnotesize
  \setlength{\tabcolsep}{2.5pt}
  \renewcommand{\arraystretch}{0.94}

  \noindent\textbf{\textcolor{taskpoolrtltext}{RTL task families}}\hfill
  \textbf{34,972 (63.1\%)}\par\vspace{1pt}
  \begin{tabular}{@{}>{\raggedright\arraybackslash}p{0.67\linewidth}
                      >{\raggedleft\arraybackslash}p{0.28\linewidth}@{}}
    \toprule
    \textbf{Functional family} & \textbf{Tasks (share)} \\
    \midrule
    Module integration \& control & 7,724 (22.1\%) \\
    Sequential state \& counters & 6,362 (18.2\%) \\
    Combinational logic \& coding & 5,341 (15.3\%) \\
    Processor \& pipeline logic & 3,339 (9.5\%) \\
    Arithmetic \& numeric datapaths & 3,172 (9.1\%) \\
    FSM \& control logic & 2,186 (6.3\%) \\
    Other RTL & 1,902 (5.4\%) \\
    Timing, clock \& reset & 1,900 (5.4\%) \\
    Memory \& buffering & 1,433 (4.1\%) \\
    Communication \& peripherals & 805 (2.3\%) \\
    Cryptography \& error control & 536 (1.5\%) \\
    DSP \& signal processing & 272 (0.8\%) \\
    \bottomrule
  \end{tabular}

  \vspace{5pt}
  \noindent\textbf{\textcolor{taskpoolkernel}{Kernel task families}}\hfill
  \textbf{20,490 (36.9\%)}\par\vspace{1pt}
  \begin{tabular}{@{}>{\raggedright\arraybackslash}p{0.67\linewidth}
                      >{\raggedleft\arraybackslash}p{0.28\linewidth}@{}}
    \toprule
    \textbf{Functional family} & \textbf{Tasks (share)} \\
    \midrule
    Matrix multiplication \& linear algebra & 3,157 (15.4\%) \\
    Elementwise math & 3,050 (14.9\%) \\
    Convolution \& vision operators & 2,893 (14.1\%) \\
    Indexing, embedding \& data movement & 1,969 (9.6\%) \\
    Normalization & 1,870 (9.1\%) \\
    Activation functions & 1,828 (8.9\%) \\
    Reductions \& statistics & 1,670 (8.2\%) \\
    Pooling \& resampling & 1,521 (7.4\%) \\
    Losses \& training objectives & 1,326 (6.5\%) \\
    Tensor creation, casting \& layout & 395 (1.9\%) \\
    Other kernel & 377 (1.8\%) \\
    Attention \& sequence modeling & 362 (1.8\%) \\
    Spectral \& signal transforms & 38 (0.2\%) \\
    Optimization \& parameter updates & 21 (0.1\%) \\
    Randomization \& sampling & 11 (0.1\%) \\
    Sparse \& graph operations & 2 (0.0\%) \\
    \bottomrule
  \end{tabular}
\end{minipage}

  \caption{\textbf{Distribution of the shared executable task pool.} The left
  chart shows the RTL--kernel split and the functional-family composition within
  each domain; the bars below compare cumulative coverage by the largest one,
  three, and five families. The right tables report the corresponding task
  counts and within-domain shares.}
  \label{fig:task-pool-distribution}
\end{figure}

The shared executable task pool contains 55,462 verifier-backed tasks and
preserves the domain and functional-family distribution of the SFT corpus. It
comprises 34,972 RTL tasks (63.1\%) and 20,490 GPU-kernel tasks (36.9\%).
Figure~\ref{fig:task-pool-distribution} reports the corresponding functional
composition.

RTL coverage centers on module integration and controllerization (22.1\%),
sequential state and counters (18.2\%), and combinational logic and coding
(15.3\%). The kernel partition is led by matrix multiplication and linear
algebra (15.4\%), elementwise mathematics (14.9\%), and convolution and vision
operators (14.1\%). It also includes losses and training objectives (6.5\%) and
attention and sequence modeling (1.8\%), alongside normalization, activation,
reduction, pooling, indexing, and data-movement operators.

\FloatBarrier

\subsection{Evaluation Infrastructure}
\label{app:evaluation-infrastructure}

Table~\ref{tab:evaluation-infrastructure} summarizes the task inventory,
generation budget, sampling protocol, reported metrics, and verification
backend used for each benchmark. For any sampling budget $k$, avg@$k$ is the
mean binary success indicator, and pass@$k$ indicates at least one successful
output among the $k$ samples.
Syntax measures compilation or elaboration success, whereas functional
correctness requires all testbenches or numerical comparisons to pass.

\begin{table}[H]
    \centering
    \caption{Evaluation infrastructure and benchmark-specific settings.}
    \label{tab:evaluation-infrastructure}
    \setlength{\tabcolsep}{3.5pt}
    \renewcommand{\arraystretch}{1.15}
    \resizebox{\linewidth}{!}{
    \begin{tabular}{
        >{\raggedright\arraybackslash}p{2.8cm}
        >{\centering\arraybackslash}p{1.0cm}
        >{\centering\arraybackslash}p{1.6cm}
        >{\raggedright\arraybackslash}p{3.5cm}
        >{\raggedright\arraybackslash}p{3.7cm}
        >{\raggedright\arraybackslash}p{3.2cm}}
        \toprule
        \textbf{Benchmark} & \textbf{Tasks} & \makecell[l]{\textbf{Max}\\\textbf{tokens}} &
        \textbf{Sampling ($n$)} & \textbf{Reported metrics} &
        \textbf{Verification} \\
        \midrule
        VerilogEval Spec-to-RTL
        & 156 & 49,152 & $n=4$, temp.\ 0.8
        & Spec-to-RTL avg@4 & Icarus Verilog simulation \\

        VerilogEval Code-complete
        & 156 & 49,152 & $n=4$, temp.\ 0.8
        & Code-complete avg@4 & Icarus Verilog simulation \\

        RTLLM
        & 50 & 49,152 & $n=4$, temp.\ 0.8
        & Functional avg@4 & Icarus Verilog simulation \\

        CVDP (cid003)
        & 78 & 49,152 & $n=5$, temp.\ 0.8
        & Functional avg@5 & Icarus Verilog + cocotb \\

        ArchXBench v1.5
        & 71 & 49,152 & $n=1$, temp.\ 0
        & Functional pass@1 & Self-checking testbenches \\

        RealBench-Module
        & 60 & 49,152 & $n=5$, temp.\ 0.8
        & Syntax/functional pass@5 & Verilator build and simulation\\

        KernelBench
        & 250
        & 58,000 & $n=1$, temp.\ 0
        & Compiled/correct/fast at pass@1
        & Numerical comparison\\

        TritonBench-G
        & 184 & 58,000 & $n=1$, temp.\ 0, zero-shot
        & Correctness pass@1 & Numerical comparison\\
        \bottomrule
    \end{tabular}
    }
\end{table}

Generation uses a single-GPU backend with tensor parallelism one, bfloat16
precision, and a GPU-memory utilization target of 0.85. We use vLLM V1 with
\texttt{VLLM\_USE\_DEEP\_GEMM=0}, the FlashInfer GDN prefill backend, and a
default maximum model length of 65,536 tokens. Experiments run on six nodes
with eight NVIDIA H20 GPUs per node. During generation, each model occupies one
node; evaluation is sharded across 40 GPUs. KernelBench uses Triton and CUDA
backends. Its L1, L2, and L3 subsets contain 100 single-operator tasks, 100
operator-fusion tasks, and 50 full-network tasks, respectively. The fast metric
counts correct kernels that also satisfy the configured speedup threshold
($1.05\times$).

Verification uses the simulator version each official harness specifies: Icarus Verilog 11.0 for ArchXBench, 12.0 for VerilogEval and RTLLM, and 13.0 for CVDP, since the versions differ materially in SystemVerilog support. RealBench-Module uses Verilator 5.049; syntax requires the build to complete without emitting any warning and functional requires a mismatch-free simulation transcript, so a build that warns scores zero on both. ArchXBench levels 0--3 are judged by their self-checking testbenches, while levels 4--6 compare against Makefile-driven golden models. Kernel benchmarks execute under PyTorch 2.9.1 with Triton 3.5.1 and CUDA 13.0. A candidate is run on the same inputs as the reference implementation and counts as correct when every returned tensor matches the reference in shape and within a tolerance of $10^{-2}$.

The GPU and software configuration above applies to the seven main benchmarks.
The separately supplied iterative kernel case-study archive does not record its
GPU model or software versions, so we do not transfer this configuration to
that analysis.

\subsection{Kernel Case-Study Operators}
\label{app:kernel-case-study-operators}

The full archive contains eleven operators: seven in-house tasks organized
under an RTLScout-inspired iterative-search scaffold
\citep{arnold2026rtlscout} and four KernelFactory samples. The main-text
visualization retains eight of them---five in-house tasks and three
KernelFactory samples. This provenance describes the task scaffolds; none of
the in-house operators is presented as an original RTLScout benchmark item.
Each task's \texttt{reference.py} designates a trusted implementation through
\texttt{baseline()} or \texttt{ref\_kernel()}. The latency values in
Table~\ref{tab:kernel-case-study-operators} are session measurements supplied
with the operator manifest, not hardware-independent constants.

\begin{table}[H]
    \centering
    \caption{Operator definitions, task provenance, and archive-designated
    reference implementations for the full eleven-task case-study archive.
    The main-text display retains eight tasks under its stated post-hoc rule.}
    \label{tab:kernel-case-study-operators}
    \small
    \setlength{\tabcolsep}{3.2pt}
    \renewcommand{\arraystretch}{1.22}
    \begin{tabular}{@{}
        >{\raggedright\arraybackslash}p{0.15\linewidth}
        >{\raggedright\arraybackslash}p{0.22\linewidth}
        >{\raggedright\arraybackslash}p{0.14\linewidth}
        >{\raggedright\arraybackslash}p{0.27\linewidth}
        >{\raggedright\arraybackslash}p{0.14\linewidth}@{}}
        \toprule
        \textbf{Operator} & \textbf{Computation} & \textbf{Task origin} &
        \textbf{Reference implementation (archived latency)} &
        \textbf{Main display} \\
        \midrule
        Vector Sum
        & Full reduction over all vector elements
        & In-house
        & \texttt{torch.sum} ($135.8\,\mu$s)
        & Yes \\

        Cumulative Sum
        & Row-wise prefix sum along dimension 1
        & In-house
        & \texttt{torch.cumsum} ($333.6\,\mu$s)
        & Yes \\

        Causal Conv1D
        & Left-padded convolution for sequence modeling
        & In-house
        & PyTorch \texttt{conv1d} with left padding ($142.9\,\mu$s)
        & Yes \\

        Asymmetric MatMul
        & Irregular-shape matrix multiplication
        & In-house
        & PyTorch matmul / cuBLAS ($1225.0\,\mu$s)
        & Yes \\

        Grayscale Conversion
        & Weighted RGB-to-luminance transform
        & In-house
        & PyTorch weighted tensor sum ($474.1\,\mu$s)
        & Yes \\

        GDNO
        & Gated Delta-Net block recurrence for linear attention
        & In-house
        & PyTorch chunked recurrence ($198.4\,\mu$s)
        & No: HY3 missing \\

        Histogram
        & Binned counting with write contention
        & In-house
        & \texttt{torch.histc}/\texttt{bincount} ($217.5\,\mu$s)
        & No: lower $r_t$ \\

        Softmax
        & Numerically stable row-wise softmax
        & KernelFactory S03
        & ATen \texttt{F.softmax}, FP32 accumulation ($903.8\,\mu$s)
        & No: lower $r_t$ \\

        SwiGLU
        & $\operatorname{SiLU}(\mathrm{gate})\odot\mathrm{up}$ activation
        & KernelFactory S04
        & \texttt{torch.compile} fused expression ($132.0\,\mu$s)
        & Yes \\

        KL Divergence
        & Temperature-scaled student--teacher distillation loss
        & KernelFactory S12
        & \texttt{torch.compile} token-blocked loss ($997.0\,\mu$s)
        & Yes \\

        Fused Cross-Entropy
        & Logits-to-loss fusion with backward-gradient readout
        & KernelFactory S11
        & Liger fused cross-entropy, with PyTorch fallback
          ($1338.0\,\mu$s)
        & Yes \\
        \bottomrule
    \end{tabular}
\end{table}

\clearpage
\subsection{Executable Verification Across Post-Training}
\label{app:verifier-design}

\paratitle{Scope-ordered contracts behind a common evidence model.}
SFT, OPSD, and RLVR reuse task-native execution contracts through
stage-specific adapters. For a generated artifact and its trusted task
context, the adapter returns an evidence record with four named fields: status
(pass/fail/unjudged), failure stage, task-grounded
measurements such as mismatch rate, assertion coverage, or speedup, and
provenance such as verifier versions and payload digests. The admissible oracle
remains local to the artifact and its trusted harness.

Tables~\ref{tab:verifier-design-rtl} and
\ref{tab:verifier-design-kernel} organize the implemented contracts by
generated-artifact boundary. The RTL
contracts form a partial scope order from module-local completion, through a
standalone module, to a module inserted into a protected project hierarchy.
Oracle choice is orthogonal to this order: at the standalone-module scope,
correctness may be established by differential traces, an explicit
self-checking verdict, programmatic assertions, or a generated-golden
pipeline. The largest generated RTL artifact in the current implementation is
a module. Project-integrated verification tests that module in a larger
dependency context.

\begin{table}[H]
    \centering
    \caption{Implemented RTL verification contracts, ordered by the boundary of
    the generated artifact. The oracle variants at the standalone-module scope
    are alternatives selected by the task specification; they do not represent
    successively larger design scales.}
    \label{tab:verifier-design-rtl}
    \footnotesize
    \setlength{\tabcolsep}{3.2pt}
    \renewcommand{\arraystretch}{1.14}
    \begin{tabular}{@{}
        >{\raggedright\arraybackslash}p{0.16\linewidth}
        >{\raggedright\arraybackslash}p{0.27\linewidth}
        >{\raggedright\arraybackslash}p{0.20\linewidth}
        >{\raggedright\arraybackslash}p{0.31\linewidth}@{}}
        \toprule
        \textbf{Verification scope} & \textbf{Candidate and trusted context} &
        \textbf{Native oracle} & \textbf{Required execution evidence} \\
        \midrule
        Module-local completion
        & Candidate completion joined to a trusted module prefix and executed
          by a fixed harness
        & Pinned Icarus transcript
        & Exactly one target module compiles and executes; the transcript
          matches its trusted profile. Mismatch-reporting tasks additionally
          require zero mismatches, a positive sample count, and no timeout. \\
        \midrule
        \multirow{4}{=}{Standalone module}
        & Complete candidate module evaluated beside trusted reference RTL
        & Reference--candidate trace comparison
        & Both designs compile under the same generated stimulus, and their
          output traces agree exactly. \\
        \cdashline{2-4}
        & Complete candidate module with a trusted self-checking harness
        & Explicit testbench verdict
        & Simulation reaches a positive verdict and reports no failure;
          missing positive evidence fails closed. \\
        \cdashline{2-4}
        & Complete candidate module embedded in an allowlisted
          Icarus/cocotb/pytest harness
        & Programmatic assertions
        & Every task-defined assertion completes and the test process exits
          successfully. \\
        \cdashline{2-4}
        & Complete candidate module with task stimuli, reference-generation
          scripts, and an output comparator
        & Task-native golden pipeline
        & The Makefile-driven generation, compilation, execution, and
          comparison pipeline completes successfully. \\
        \midrule
        Project-integrated module
        & Complete candidate module inserted into a protected dependency
          hierarchy with a trusted reference instance and stimulus generator
        & Differential project simulation
        & The Verilator project builds and runs; its normalized transcript
          matches the pinned trusted profile. \\
        \bottomrule
    \end{tabular}
\end{table}

\begin{table}[H]
    \centering
    \caption{Implemented GPU verification contracts. Both compare generated
    code with trusted reference execution under task-defined inputs, while the
    admissible implementation boundary and efficiency supervision differ.}
    \label{tab:verifier-design-kernel}
    \footnotesize
    \setlength{\tabcolsep}{3.4pt}
    \renewcommand{\arraystretch}{1.16}
    \begin{tabular}{@{}
        >{\raggedright\arraybackslash}p{0.18\linewidth}
        >{\raggedright\arraybackslash}p{0.29\linewidth}
        >{\raggedright\arraybackslash}p{0.20\linewidth}
        >{\raggedright\arraybackslash}p{0.27\linewidth}@{}}
        \toprule
        \textbf{Artifact boundary} & \textbf{Candidate and trusted context} &
        \textbf{Native oracle} & \textbf{Accepted evidence and measurement} \\
        \midrule
        Operator implementation
        & Candidate device-level operator and a trusted executable reference,
          run with matched deterministic seeds
        & Structured numerical equivalence
        & Output structures and shapes agree and tensor values satisfy the
          task tolerance. A completed device launch is required; prohibited
          framework delegation and identity stores invalidate the result. \\
        \midrule
        Reference-model replacement
        & Candidate replacement module, trusted reference model, and
          task-defined initialization and input generators
        & Repeated output equivalence under matched inputs
        & Correctness is checked across repeated trials. Matched timing against
          the reference supplies a separate speedup measurement; unchanged
          framework operators outside the replacement remain admissible. \\
        \bottomrule
    \end{tabular}
\end{table}

\paratitle{Stage-specific use of executable evidence.}
The three post-training stages apply these contracts to different decisions.
SFT uses the correctness verdict as an admission gate. OPSD preserves execution
stages and task-native measurements as feedback and failure-conditioned
experience. RLVR maps verified outcomes and eligibility signals to scalar
rewards.

\paratitle{Trusted harnesses and candidate isolation.}
Verifier inputs are divided into trusted task material and untrusted model
output. Serialized payloads are size bounded, schema checked, and restricted to
relative paths. Depending on the backend, trusted task material is bound by
SHA-256 allowlists, task tokens, or verdict profiles; the profiles pin a dataset
revision, toolchain version, and expected transcript digest. The online service
also fingerprints its verifier modules, GPU evaluator, and trusted task source
tree at startup. A missing dependency, profile mismatch, or source-tree change
therefore returns an unverified status. RTL
candidates are extracted separately, screened for protected-module
collisions and unsafe runtime constructs, and compiled or executed in a
temporary directory under time, memory, file, and process limits. GPU requests
receive a fresh process group and CUDA context; their results cross a bounded
JSON channel whose fields and types are canonicalized before the service
accepts them.

\paratitle{Execution evidence and anti-exploitation checks.}
For GPU code, a runtime launch-exit hook counts completed Triton launches;
compilation, warm-up, and kernel-object indexing do not satisfy the execution
requirement. Static checks additionally flag stores that
return loaded values unchanged and framework operations that retain prohibited
computation outside Triton. These checks are scoped by the generated-artifact
contract. Operator-only contracts forbid framework delegation of the requested
computation, whereas replacement contracts permit unchanged framework
operators outside the replaced computation and measure efficiency against the
reference execution. The verifier exports these signals separately from
correctness, and the reward adapter applies the relevant signals as eligibility
gates before assigning the kernel reward in
Eq.~\eqref{eq:rlvr-kernel-reward}.

\paratitle{Failure attribution and RLVR reward construction.}
Only allowlisted failures from a trusted control-plane channel are marked
unverified; candidate output cannot trigger this state. After bounded retries,
an unresolved RLVR sample receives the sentinel $-1$ and is removed from group
statistics and policy updates (Eq.~\eqref{eq:rlvr-masked-advantage}). SFT and
OPSD retain their original verdict and diagnostic semantics. RTL rewards follow
Eq.~\eqref{eq:rlvr-rtl-reward}. GRPO compares responses to the same prompt, and
the domain sampling mixture controls each task family's update frequency.

\end{document}